\documentclass[letterpaper]{article} 
\usepackage{aaai2026}  
\usepackage{times}  
\usepackage{helvet}  
\usepackage{courier}  
\usepackage[hyphens]{url}  
\usepackage{graphicx} 
\usepackage{natbib}  
\usepackage{caption} 
\usepackage{algorithm}
\usepackage{algorithmic}
\usepackage{multirow}
\usepackage{graphicx}
\usepackage{dblfloatfix}
\usepackage{tcolorbox}
\usepackage{enumitem}
\usepackage{tabularx}
\usepackage{subcaption} 
\usepackage{adjustbox}
\usepackage{booktabs}
\usepackage{siunitx} 
\usepackage{caption}
\usepackage{xcolor}
\usepackage{multirow}
\usepackage[table]{xcolor} 
\usepackage{tabularx}
\usepackage{booktabs}
\newcolumntype{Y}{>{\raggedright\arraybackslash}X}

\usepackage{xcolor}

\newcommand{\answerTODO}[1]{\textcolor{red}{#1}}
\usepackage{newfloat}
\usepackage{listings}
\DeclareCaptionStyle{ruled}{labelfont=normalfont,labelsep=colon,strut=off} 
\floatstyle{ruled}
\newfloat{listing}{tb}{lst}{}
\floatname{listing}{Listing}
\title{Reddit After Roe: A Computational Analysis of Abortion Narratives and Barriers in the Wake of Dobbs}
\author {
    Aria Pessianzadeh,\textsuperscript{\rm 1}
    Alex H. Poole, \textsuperscript{\rm 1}
    Rezvaneh Rezapour \textsuperscript{\rm 1}
}
\affiliations {
    \textsuperscript{\rm 1} Information Science Department, Drexel University, United States \\
    \{ap3943, ahp56, shadi.rezapour\}@drexel.edu
}

\begin{document}
\maketitle

\begin{abstract}
The 2022 U.S. Supreme Court decision in Dobbs v. Jackson Women's Health Organization reshaped the reproductive rights landscape, introducing new uncertainty and barriers to abortion access. We present a large-scale computational analysis of abortion discourse on Reddit, examining how barriers to access are articulated across information-seeking and information-sharing behaviors, different stages of abortion (before, during, after), and three phases of the Dobbs decision in 2022.
Drawing on more than 17,000 posts from four abortion-related subreddits, we employed a multi-step pipeline to classify posts by information type, abortion stage, barrier category, and expressed emotions. Using a codebook of eight barrier types, including legal, financial, emotional, and social obstacles, we analyzed their associations with emotions and information behaviors. Topic modeling of model-generated barrier rationales further revealed how discourse evolved in response to shifting legal and cultural contexts.
Our findings show that emotional and psychological barriers consistently dominate abortion narratives online, with emotions such as nervousness, confusion, fear, and sadness prevalent across discourse. By linking information behaviors, barriers, emotions, and temporal dynamics, this study provides a multi-dimensional account of how abortion is navigated in online communities.
\end{abstract}

\begin{links}
    \link{Code and Dataset}{https://github.com/social-nlp-lab/Dobbs-Abbortion-Narratives}

\end{links}

\section{Introduction}\label{sec:introduction}

The U.S. Supreme Court's decision in Dobbs v. Jackson Women's Health Organization (2022) marked a pivotal moment in the national landscape of reproductive rights~\cite{chang2023roeoverturned}.
In its aftermath, individuals increasingly turned to social media to express opinions, share personal experiences, and seek guidance amid rapidly changing legal conditions. As traditional pathways to reproductive healthcare become restricted and challenging to access, online communities provide an accessible venue for those seeking care to navigate uncertainty, locate resources, and articulate the personal and political stakes of abortion \cite{dolgins2025behavioral, wilson2024seeking}.
These platforms simultaneously reflect public attitudes and function as infrastructures for information exchange, making them critical sites for large-scale computational analysis of abortion discourse \cite{neely2021health, chen2021social}. 

Abortion access has long been shaped by multiple intersecting barriers, ranging from restrictive legislation to financial, logistical, and social constraints \cite{jerman2017barriers, pleasants2024abortion}. 
Decades of research demonstrate that cost, distance to clinics, inadequate provider availability, and stigma consistently limit access to abortion care, often disproportionately affecting marginalized populations \cite{gerber1997abortion, doran2015barriers}. These barriers not only delay or prevent care but also exacerbate emotional and psychological burdens for those seeking services \cite{pleasants2024normal}. The Dobbs decision has further intensified these constraints, increasing uncertainty and complexity in navigating reproductive healthcare. We examine how these barriers are expressed and discussed in Reddit communities, where users actively seek and share information in response to shifting legal and social landscapes.

While prior research has documented various types of abortion barriers, existing studies largely rely on small-scale qualitative analyses and often examine barriers in isolation, without systematically linking them to emotional expression or information behavior~\cite{pleasants2024abortion, jerman2017barriers, higgins2021real, lee2023barriers}. We address this gap through a large-scale computational analysis of Reddit discussions that jointly models abortion-related barriers, emotional expression, and information behaviors.
This integrated perspective provides a more comprehensive understanding of how structural constraints and lived experiences are intertwined in online abortion discourse. We answer the following research questions:

\begin{itemize}[leftmargin=*,noitemsep]

\item RQ1: How do information behaviors shape abortion discourse on Reddit across different phases of the Dobbs decision and stages of abortion? 

\item RQ2: What barriers to abortion are discussed in Reddit narratives, and how do these barriers intersect with information behaviors and phases of the Dobbs decision?

\item RQ3: What emotional expressions are present in abortion-related discourse, and how do they vary across information behaviors and barrier categories?

\end{itemize}

To address these questions, we collected data from four Reddit communities (N = 17,534) that primarily focus on abortion-related discussions. We analyze how users seek and share information while articulating barriers across abortion stages (before, during, and after) and phases of the Dobbs decision. 
By linking information behaviors with barrier types, emotional expressions, and thematic patterns, our framework captures how abortion is experienced, constrained, and communicated in online spaces.
Our findings show that information-seeking behavior substantially outnumbers information-sharing, highlighting Reddit's primary role as a space for seeking guidance and support. 
\textit{Emotional \& Psychological} barriers are the most prevalent across narratives, while \textit{Informational \& System Navigation} and \textit{Medical \& Physical} are most closely associated with information-seeking contexts and during-abortion experiences.
Further statistical analyses showed that barrier and emotion distributions vary most strongly by information behavior and abortion stage, and remain comparatively stable across Dobbs phases. 
The emotion analysis showed that \textit{nervousness}, \textit{confusion}, and \textit{fear} tend to dominate narratives of abortion, indicating the emotionally loaded nature of abortion discourse.

This work advances research on abortion discourse by providing a computational analysis of information-seeking and information-sharing behaviors, while linking barriers, emotions, and information types. By situating these dynamics across the three phases of the Dobbs decision, it offers new insights into how individuals navigate abortion in the digital public sphere.

\section{Related Work}
\label{sec:relatedwork}

\paragraph{Abortion Discourse on Social Media Platforms.}
Social media platforms have become central to abortion discourse, particularly in the wake of the recent Dobbs decision that overturned Roe v. Wade \cite{dolgins2025behavioral, wilson2024seeking}.
Reddit, with its community-based structure and potential to post longer messages, fosters both debate and support \cite{de2014mental, zhang2017community}. \citet{valdez2024analyzing} compared \textit{r/abortion} and \textit{r/abortiondebate}, showing that the former emphasized support while the latter emphasized legal and moral arguments. \citet{stanier2024polarization} added nuance by showing how frames of bodily autonomy, legality, and morality diverged across subreddits. 
In other platforms, \citet{aleksandric2024analyzing} used stance detection on Facebook to analyze posts across states, finding that abortion attitudes online often mirrored state-level political and health indicators. On Twitter/X, studies revealed how polarization, framing, and engagement intersect. \citet{philippe2024abortion} showed that polarity and subjectivity shape engagement patterns, while \citet{dai2024social} identified influential actors and hashtag networks around Roe's reversal. \citet{mane2022examination} observed geographic divides in sentiment, with pro-life content concentrated regionally. Large-scale datasets provided further context; \citet{chang2023roeoverturned} compiled 74 million tweets on Roe, and \citet{rao2025polarized} demonstrated how court decisions triggered spikes in antagonistic frames. Earlier work also showed how gender and stance shape abortion discourse \cite{durmus2018understanding}. 
\citet{fredenburg2023camping} showed how euphemisms such as ``camping'' function as coded language for discussing abortion in restrictive contexts, reflecting how online communities adapt to stigma and surveillance. \citet{hanschmidt2016abortion} highlighted stigma as a central social frame, with abortion often narrated through shame, secrecy, and moral judgment. Complementing this, work grounded in Moral Foundations Theory highlighted how ideological divides over abortion map onto broader cultural values such as purity, harm, and fairness \cite{sharma2017analyzing, rezapour-etal-2019-enhancing, rezapour2021incorporating}.

Using temporal analyses, \citet{venkata2024post} found that abortion discourse on Reddit and YouTube transitioned from ethical frames immediately post-Dobbs, toward legal and political frames during peak debates, and eventually back to ethical narratives in later stages. \citet{cowan2024updating} showed that abortion attitudes in the U.S. remain dynamic, with shifting public consensus that aligns with discourse changes observed online. \citet{jacques2021medication} further demonstrated how decision-making and framing on social media reflect both personal experience and broader cultural debates. Other studies highlight privacy and trust concerns, documenting fears around surveillance, digital security, and institutional legitimacy \cite{guo2024perspectives, song2024our}

\paragraph{Information Seeking, Sharing, and Support in Online Communities. }
Abortion discourse on social media transcends debate or polarization; it also reflects the various ways individuals turn to online communities for information and mutual support \cite{rezapour2023moving}. 
\citet{john2024abortion} identified informational and emotional needs expressed in \textit{r/abortion} posts following Dobbs, showing that users often sought clarity about medical procedures, legal or logistical arrangements, while also voicing fear, stigma, and uncertainty.
\citet{pleasants2024normal} demonstrated that Reddit posts often combine emotional disclosure with concrete informational requests.
Reddit is also where people share their personal lived experiences \cite{bouzoubaa2024decoding}. \citet{pleasants2024waiting} showed how users provided guidance on managing waiting times and overcoming logistical hurdles. 
\citet{dolgins2025behavioral} highlighted community-driven innovations such as self-management strategies, privacy-oriented practices, and funding solutions that circulated in online exchanges.
Other studies examined discourse around self-managed abortion \cite{weidert2025have}, insurance restrictions and healthcare costs \cite{higgins2021real}, and the role of social media platforms as infrastructures for individuals navigating reproductive health \cite{moseson2022just}

\paragraph{Barriers to Abortion Access.}
\citet{pleasants2024abortion} documented how users described state bans, gestational limits, and stigma as primary barriers, often linking these to emotional distress and delayed access to care. 
\citet{doran2015barriers} reviewed first-trimester abortion obstacles, identifying distance, cost, and stigma as consistent challenges. \citet{culwell2013addressing} outlined barriers to abortion, emphasizing the intersection of law, policy, and provider-level issues. 
Other work highlighted the persistence and adaptation of barriers across contexts. \citet{jacques2023m} analyzed Reddit during COVID-19, showing how pandemic disruptions compounded existing challenges. \citet{lee2023barriers} focused on provider-side barriers, including inadequate training and institutional opposition, while \citet{bernstein2023practice} found that abortion restrictions influenced physicians' and trainees' practice location choices. 

Our study builds on prior research on abortion discourse, information exchange, and access barriers on social media, which are often examined separately. We address this gap by analyzing Reddit discussions across phases of the Dobbs decision to understand how information behaviors, barriers, and emotional expression intersect and evolve.

\section{Method}
\label{sec:method}

\subsection{Data Collection}
We used Pushshift Reddit API \cite{baumgartner2020pushshift} to collect data from four abortion-related subreddits: \textit{r/abortion}, \textit{r/abortiondebate}, \textit{r/prochoice}, and \textit{r/prolife}, that discuss abortion from different perspectives. 
Pre-processing steps were done to remove deleted posts (tagged as ['deleted']) or posts without a text body.
As our primary focus is on how abortion discourse is shaped by news related to Roe v. Wade and its implications, we filtered the dataset to include only posts from six months before and six months after the official overturning on June 24, 2022, resulting in 17,534 posts for analysis. To capture shifts in discourse, we further divided the 2022 timeline into three phases: Phase 1 (January 1–May 1) represents the period before Dobbs; Phase 2 (May 1–September 1) covers the months immediately surrounding the decision; and Phase 3 (September 1–December 31) reflects the period after the ruling. These phases capture how discussions of abortion barriers shifted following the Supreme Court decision.

\subsection{Narrative Classification and Analysis}

\paragraph{Information Behaviors: Seeking vs. Sharing. }
The distinction between information seeking and sharing has been studied in communication and information science \cite{poltrock2003information}. In health contexts, information seeking involves asking for advice, clarification, or resources to navigate uncertainty, while information sharing encompasses the disclosure of personal experiences, narratives, or advice to others \cite{savolainen2005everyday}. These two behaviors are mutually reinforcing in online communities, where users simultaneously contribute knowledge and rely on peer support \cite{goodyear2019young}. Within abortion discourse, distinguishing between these two functions helps examine how individuals mobilize online spaces both as infrastructures of information exchange and as venues of support-seeking and storytelling. This dimension is particularly salient on Reddit, whose community-based structure encourages both forms of communication \cite{de2014mental}.

To operationalize these dimensions, we manually annotated 400 Reddit posts. Two independent annotators labeled each post for information behavior (seeking vs.\ sharing), with disagreements resolved through discussion and, when necessary, adjudicated by a third annotator. Inter-annotator reliability was high (Cohen's $\kappa = 0.89$). 
Using a subset of this annotated data, we trained and evaluated multiple classification models, including transformer-based models (BERT, RoBERTa, DistilBERT, DeBERTa) and state-of-the-art large language models (LLMs) (GPT-4.1, GPT-4.1-mini, GPT-5-mini, among others). The dataset was split into training, validation, and test sets, comprising 240, 60, and 100 posts, respectively. Model performance was assessed against the annotated test set using precision, recall, and F1 scores (See Appendix for the prompts). 

\paragraph{Temporal Stages of Abortion. }
Prior research showed that abortion barriers vary across different stages of the procedure \cite{jerman2017barriers}. Narratives before an abortion often emphasize logistical challenges such as travel, scheduling, and financial constraints, whereas narratives during the procedure more frequently reflect uncertainty, fear, or concerns about complications. In contrast, post-abortion narratives tend to focus on recovery, stigma, and emotional reflection \cite{kimport2011social}. To capture these stage-specific dynamics, we classified posts into five categories, \textit{Before}, \textit{During}, \textit{After}, \textit{Not Sure}, and \textit{Irrelevant}, reflecting how experiences unfold across the abortion process.  
Two annotators independently labeled the same set of 400 posts using these categories. Disagreements were resolved through discussion, and inter-annotator reliability was assessed using Cohen's kappa ($\kappa = 0.85$), indicating strong agreement.
We evaluated four LLMs (GPT-4.1, GPT-4.1-mini, GPT-5-mini, and GPT-5-nano) for abortion stage classification using the same human-annotated ground truth set (See Appendix for the prompts).

\paragraph{Barriers to Abortion Access. }
Barriers to abortion access are multidimensional and have been documented across public health, reproductive justice, and social science research. Prior work identified a combination of structural, logistical, interpersonal, and psychological obstacles that shape how individuals navigate abortion care \cite{jerman2017barriers, doran2015barriers, hanschmidt2016abortion}. These barriers not only influence the feasibility and timing of care but also structure the emotional and informative experiences surrounding abortion decisions. To situate our framework within this established body of research, we adapted the barrier typology introduced by \citet{pleasants2024abortion}, integrating insights from reproductive health literature to ensure theoretical coherence and interpretability. We refined this framework through open coding of a subset of posts to ensure alignment with narrative patterns in our dataset. Our final codebook of barriers includes:
\begin{itemize}

\item  \textit{Legal \& Policy}: Restrictions arising from state laws, gestational limits, mandatory waiting periods, and criminalization fears, consistent with prior documentation of how governance systems limit access \cite{gerber1997abortion, jones2017characteristics}.

\item \textit{Financial \& Insurance}: Obstacles related to costs, insurance gaps, and travel expenses—well-established determinants of restricted access \cite{higgins2021real}.

\item \textit{Logistical \& Geographical}: Issues of clinic scarcity, long travel distances, or scheduling delays, echoing the centrality of geographic access found in prior research \cite{doran2015barriers, jerman2017barriers}.

\item \textit{Provider \& Infrastructure}: Constraints related to clinic capacity, provider availability, or institutional limitations, aligning with research on system-level access challenges \cite{lee2023barriers}.

\item \textit{Medical \& Physical}: Concerns about symptoms, complications, or procedural risk—barriers shown to influence decision-making and care-seeking, especially in medication or self-managed abortions \cite{upadhyay2015incidence}.

\item \textit{Informational \& System Navigation}: Difficulties understanding legal requirements, procedures, medication instructions, reflecting fragmented health-information environments \cite{neely2021health}.

\item \textit{Emotional \& Psychological}: Experiences of fear, anxiety, shame, guilt, and uncertainty, grounded in stigma research \cite{hanschmidt2016abortion, norris2011abortion}.

\item \textit{Social \& Interpersonal}: Challenges involving partner conflict, lack of support, or familial stigma, consistent with findings that interpersonal relationships shape emotional and practical aspects of abortion decisions \cite{kimport2011social}.

\end{itemize}

Posts tagged as \textit{Before}, \textit{During}, and \textit{After} were coded by two annotators for the presence of different barrier types, as these stages most directly reflect concrete abortion experiences where barriers were discussed. Since posts could reference multiple barriers, this task was treated as a multi-label classification, with each barrier type coded as an independent binary dimension.
Inter-annotator reliability was assessed using Cohen's kappa for each barrier category, with an average agreement of $\kappa = 0.64$, indicating substantial reliability. Category-specific values are reported in Table~\ref{tab:merged_barriers}. Lower kappa scores for \textit{Provider \& Infrastructure} ($\kappa = 0.28$) and \textit{Informational \& Navigation} ($\kappa = 0.30$) are attributable to class imbalance. To account for this, we report Gwet's AC1, which remains high for these categories ($0.99$ and $0.84$), indicating strong agreement. Overall, the results demonstrate high annotation reliability across categories.

\begin{table}[t]
\centering
\begin{tabular}{l c c}
\toprule
\textbf{Barrier Category} & \textbf{$\kappa$} & \textbf{AC1 } \\ \midrule
Legal \& Policy            & 0.89 & 0.99 \\
Financial \& Insurance      & 0.89 & 0.99 \\
Logistical \& Geographical  & 0.69 & 0.98 \\
Provider \& Infrastructure  & 0.28 & 0.99 \\
Medical \& Physical         & 0.57 & 0.96 \\
Informational \& Navigation  & 0.30 & 0.84 \\
Emotional \& Psychological   & 0.72 & 0.91 \\
Social \& Interpersonal     & 0.81 & 0.98 \\ \midrule
\textbf{Average Across Categories}  & \textbf{0.64} & \textbf{0.96} \\ \bottomrule
\end{tabular}
\caption{Inter-annotator agreement comparison (Cohen's $\kappa$ vs. Gwet's AC1) for the eight abortion barrier categories.}
\label{tab:merged_barriers}
\end{table}

We evaluated eight OpenAI models and two open-source models for identifying barrier types in posts, comparing their performance against our ground truth data. Performance was assessed using precision, recall, and F1 score for each barrier category, along with the average F1 score across all barrier types for each model (See Appendix for the prompts).

\paragraph{Emotion Analysis. }
Emotional expression plays an important role in abortion narratives, shaping how individuals describe experiences and engage in public discourse \cite{hanschmidt2016abortion}. To capture these expressions systematically, we used GoEmotions framework \cite{demszky2020goemotions}, which provides a taxonomy of 28 fine-grained emotions. Prior work has shown the importance of emotion detection in understanding online discourse, especially for health and political topics \cite{de2014mental, mohammad2013crowdsourcing}. By applying this framework, we identified which emotions were most salient in abortion narratives, how they differed across information behaviors and barrier categories, and how they co-occurred in complex affective narratives. 
We prompted GPT-4o (using the definitions of each category) to label each post with the most salient emotions present in the text (more information on classification performance is provided in the Appendix).

\subsection{Topic Modeling for Barriers}
We conducted a thematic analysis of posts tagged with barriers to characterize the narratives associated with each type of obstacle. To support interpretability, we leveraged chain-of-thought prompting \cite{wei2022chain}. 
We applied topic modeling to the posts to examine how barrier-related narratives varied across information behaviors (seeking vs.\ sharing) and temporal phases (Phases 1–3). For topic extraction, we employed BERTopic \cite{grootendorst2022bertopic} in combination with GPT-4.1, enabling semantically rich clustering of texts and the identification of interpretable themes.

\section{Results}
\label{sec:results}

\begin{table}[t]
        \centering
        \resizebox{0.8\columnwidth}{!}{%
        \begin{tabular}{lcccc}
        \toprule
        Model & Precision & Recall & F1-Score \\
        \midrule
            BERT   & 0.86 & 0.86 & 0.86 \\
            RoBERTa   & 0.93 & 0.93 & \textbf{0.93} \\
            DistilBERT   & 0.83 & 0.83 & 0.83 \\
            DeBERTa  & 0.88 & 0.87 & 0.87 \\
            Llama  & 0.78 & 0.75 & 0.76 \\
            GPT-4  & 0.86 & 0.85 & 0.85 \\
            GPT-4o  & 0.91 & 0.90 & 0.90\\ \hline
        \end{tabular}}
    \caption{Classification of Information-Seeking vs. Sharing using 400 annotated posts}
        \label{tab:classification_info_type}
            \vspace{-0.5cm}
\end{table}

\subsection{Data Characteristics}
\begin{figure}[t]
    \centering

    \includegraphics[width=0.8\columnwidth]{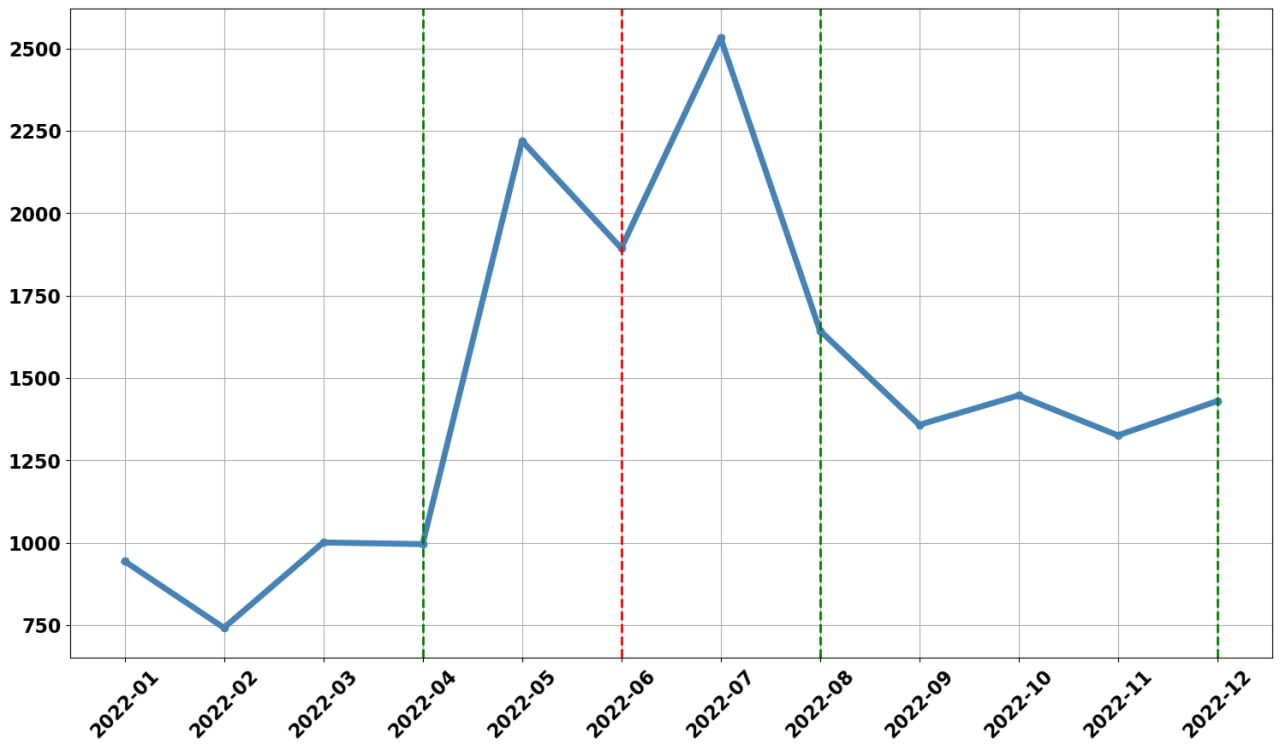}

    \caption{Number of abortion-related posts in 2022}
    \label{fig:Number of posts}
\end{figure}

Figure \ref{fig:Number of posts} shows the temporal distribution of posts across the study period. An increase in posting activity is observed at the onset of Phase 2 (May 2022), with volumes peaking in late June and early July, immediately following the Supreme Court's Dobbs decision on June 24. Although activity declined somewhat during Phase 3, posting levels remained higher than those in Phase 1, indicating a sustained elevation in discourse after the ruling.

\subsection{RQ1: Classification of Information Behaviors. }
\begin{figure}[t]
        \centering
        \includegraphics[width=0.8\columnwidth]{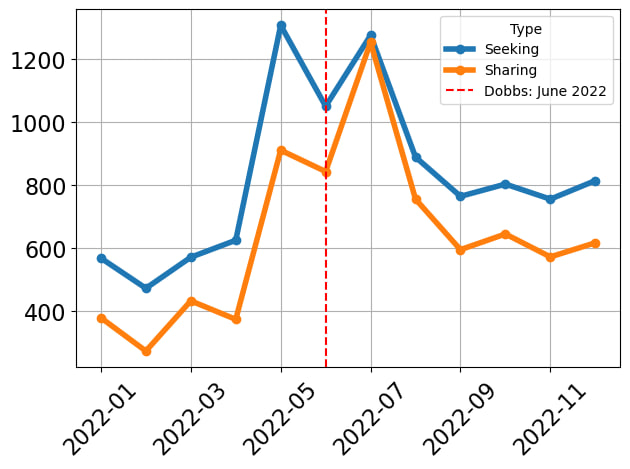}
        \caption{RoBERTa results for information types across the whole dataset}
        \label{fig:roberta-info}
            \vspace{-0.5cm}
\end{figure}

Table \ref{tab:classification_info_type} shows the performance of different models on the test set for information seeking vs.\ sharing. Fine-tuned transformer-based models performed best overall, with RoBERTa achieving the highest F1 score (0.93), outperforming LLM-based models. 
Given its strong performance, we used RoBERTa to label information types across the full dataset.
Applying this model, 9,894 posts (56.4\%) were labeled as information-seeking and 7,640 (43.6\%) as information-sharing.
Figure \ref{fig:roberta-info} shows temporal trends in posts classified as information-seeking vs.\ sharing in our dataset. Across the entire timeline, information-seeking outnumbered information-sharing, reflecting Reddit's role as a space for seeking guidance. Both categories increased sharply around the leak and subsequent announcement of the Dobbs decision (May–July 2022). Although activity declined somewhat afterward, levels of both information-seeking and sharing remained higher than in early 2022, suggesting a sustained shift in discourse following the ruling.

\begin{table*}[ht]
\centering
\begin{tabular}{c c}
    \begin{subtable}[t]{0.45\textwidth}
    \centering
    \begin{tabular}{lcc}
    \toprule
    Stage & Annotated Posts & Entire Dataset \\
    \hline
    Irrelevant & 227 & 8986 \\
    After & 84 & 3,718 \\
    Before & 62 & 3,502 \\
    During & 20 & 1,144 \\
    Not Sure & 7 & 184 \\ \hline
    
    \end{tabular}
    \caption{Number of posts with different abortion stages in manually annotated posts and the entire dataset.}
    \label{tab:stage_dist}
    \end{subtable} &

    \begin{subtable}[t]{0.45\textwidth}
    \centering
    \begin{tabular}{cccc}
    \toprule
    & & Seeking & Sharing \\
    \hline
    & P 1     & 1,415	    & 824	     \\
    Phases& P 2     & 1,945    & 1,096    \\
    & P 3     & 1,998    & 1,086    \\
    \hline
    &Before      & 2,508    & 994     \\
    Stages&During      & 772     & 372     \\
    &After       & 2,078    & 1,640    \\ \hline
    \end{tabular}
    \caption{Information labels across stage and phase.}
    \label{tab:info_stage_phase}
    \end{subtable}
\end{tabular}
\caption{Comparison of abortion-related stages and information-seeking/sharing across phases.}
\label{tab:combined_stage_info}

\end{table*}

\subsection{Temporal Stages of Abortion}

GPT-4.1 achieved the highest performance for classifying abortion stages (F1 = 0.90) compared to other models (Table~\ref{tab:stage_f1_models} in the Appendix). We therefore used this model to label abortion stages in our dataset. Table \ref{tab:stage_dist} reports the distribution of posts across abortion stages in both the annotated sample and the full dataset. The majority of posts were classified as \textit{Irrelevant}, reflecting discussions focused on ideological or political debates (e.g., pro-choice vs.\ pro-life arguments or fetal personhood) rather than personal abortion experiences. Posts tagged as \textit{After} are the most prevalent, exceeding those classified as \textit{Before} or \textit{During}. Since our research centers on narratives with lived experiences and the barriers they reveal, we restricted subsequent analyses to posts labeled as \textit{Before}, \textit{During}, and \textit{After}.

Table~\ref{tab:info_stage_phase} presents the distribution of information behaviors across Dobbs phases and abortion stages for posts labeled as \textit{Before}, \textit{During}, and \textit{After}. Across all phases, information-seeking consistently outnumbered information sharing, with noticeable increases following the Dobbs decision. Information-seeking posts increased from 1,415 in Phase~1 to 1,945 in Phase~2, peaking at 1,998 in Phase~3, reflecting heightened demand for guidance amid changing legal and logistical conditions. Information sharing followed a similar pattern, rising from 824 posts in Phase~1 to just over 1,080 posts in Phases~2 and~3, suggesting a steady reliance on Reddit for sharing experiences throughout the year. 
When comparing across abortion stages, information-seeking was most prevalent \textit{Before} seeking abortion (2,508 posts), highlighting the role of social media as a key resource for gathering advice and preparing for procedures. In contrast, information sharing was most common \textit{After} seeking abortion (1,640 posts), highlighting how users turn to Reddit to reflect on and process their experiences.

\subsection{RQ2: Barriers to Abortion Access}
Classification performance for barrier types varied across categories (see Table \ref{tab:Barriers Classification Models}): Barriers related to \textit{Legal \& Policy} and \textit{Logistical \& Geographical} factors achieved higher performance, whereas categories such as \textit{Provider \& Infrastructure} and \textit{Medical \& Physical} were more challenging to identify, likely reflecting their greater contextual and experiential complexity. Among all models evaluated, GPT-4o achieved the highest average F1 score (0.75). Open-source models, including LLaMA and Mixtral, performed substantially worse and did not achieve comparable results.
Based on overall performance, we selected the three highest-performing models, GPT-4o (F1 = 0.75), GPT-5.1 (F1 = 0.74), and GPT-4.1-mini (F1 = 0.72), to label the remaining dataset. For each post and barrier type, we assigned the final label based on majority agreement across these three models, allowing posts to be associated with one or multiple barrier categories (We did not include GPT-4 due to cost considerations).

\begin{figure*}[b]
    \centering
    \includegraphics[width=0.6\textwidth]{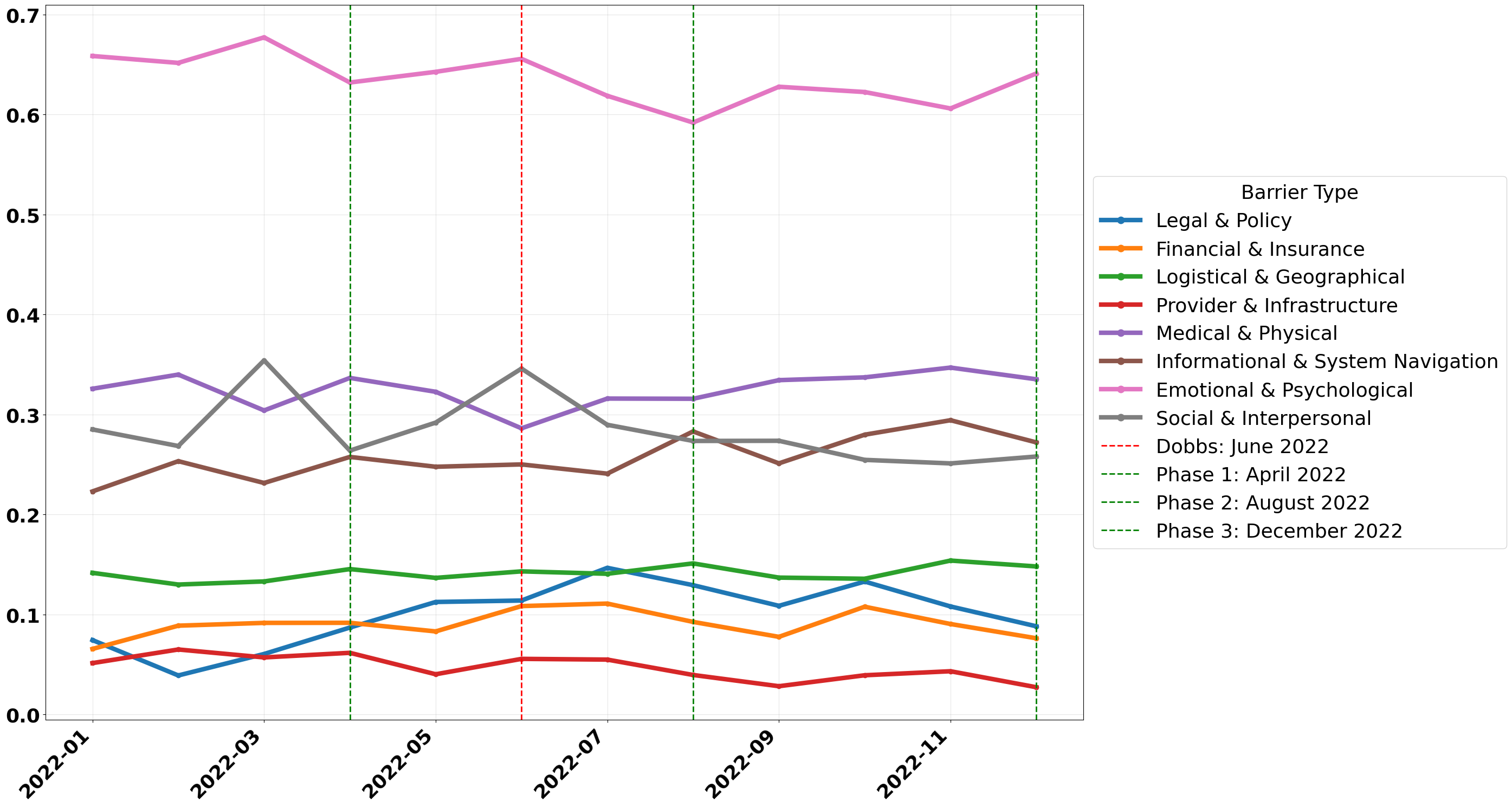}
    \caption{Temporal Changes in Different Types of Abortion Barriers}
    \label{fig:Barriers_Temporal}
\end{figure*}

Figure \ref{fig:Barriers_Temporal} shows temporal changes in the distribution of barrier types in our dataset. Mentions of \textit{Emotional \& Psychological} barriers consistently dominated the discourse, followed by \textit{Medical \& Physical} and \textit{Social \& Interpersonal} challenges. By contrast, barriers related to \textit{Provider \& Infrastructure} or \textit{Financial} constraints were less frequent, as these challenges affect a narrower subset of abortion seekers. We also observed a steady increase in the proportion of posts referencing \textit{Legal \& Policy} barriers over time, likely reflecting heightened attention to the legal ramifications of the Dobbs decision. Table \ref{tab:Barriers that co-occur} (in the Appendix) highlights the most frequent pairs of barriers that co-occur in each post.

\begin{figure}[ht]
\centering

\begin{subfigure}[t]{\columnwidth}
    \centering
    \includegraphics[width=0.8\columnwidth]{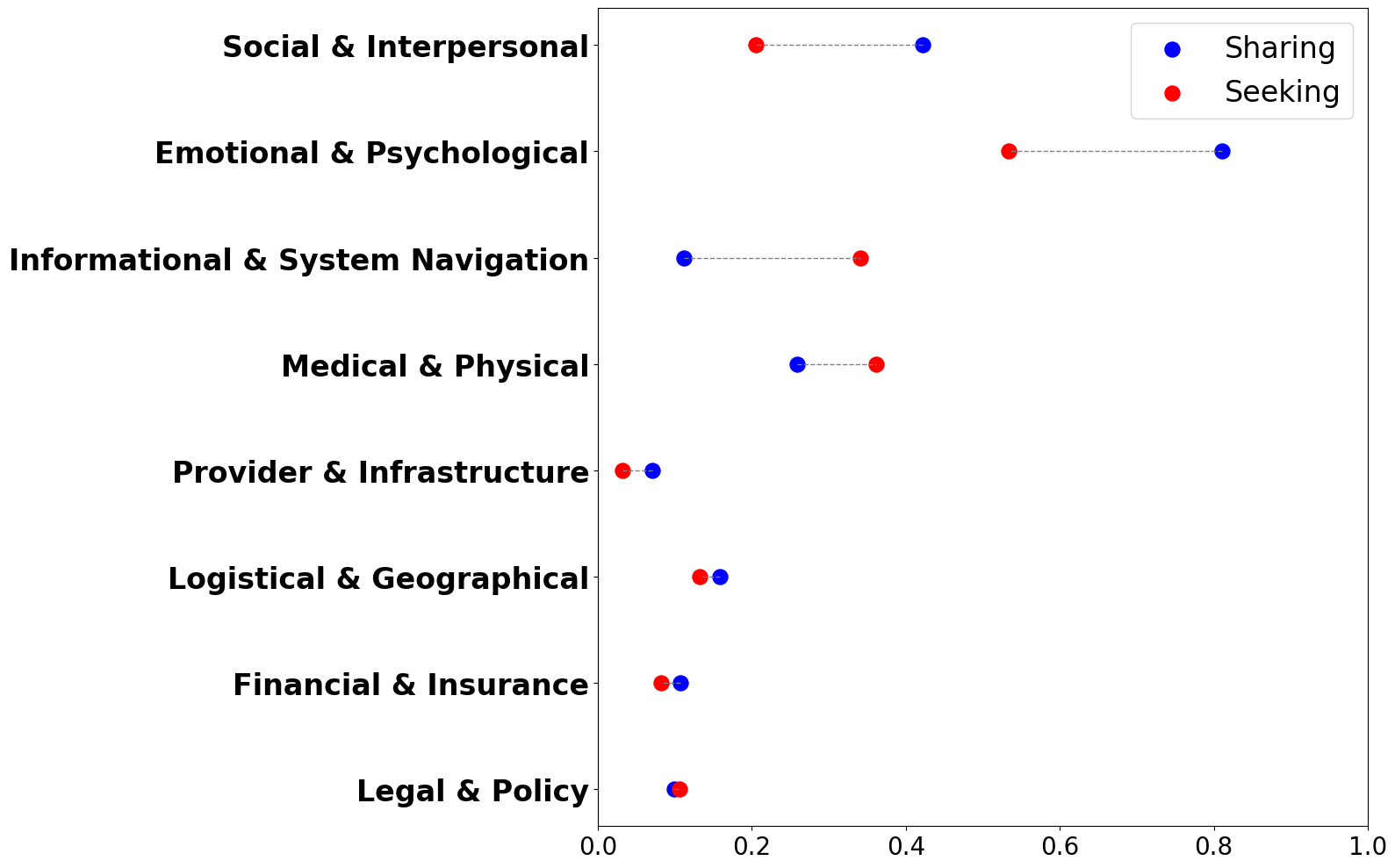}
    \caption{Barriers in two information types}
    \label{fig:Barriers_Info}
\end{subfigure}

\vspace{0.5em}

\begin{subfigure}[t]{\columnwidth}
    \centering
    \includegraphics[width=0.8\columnwidth]{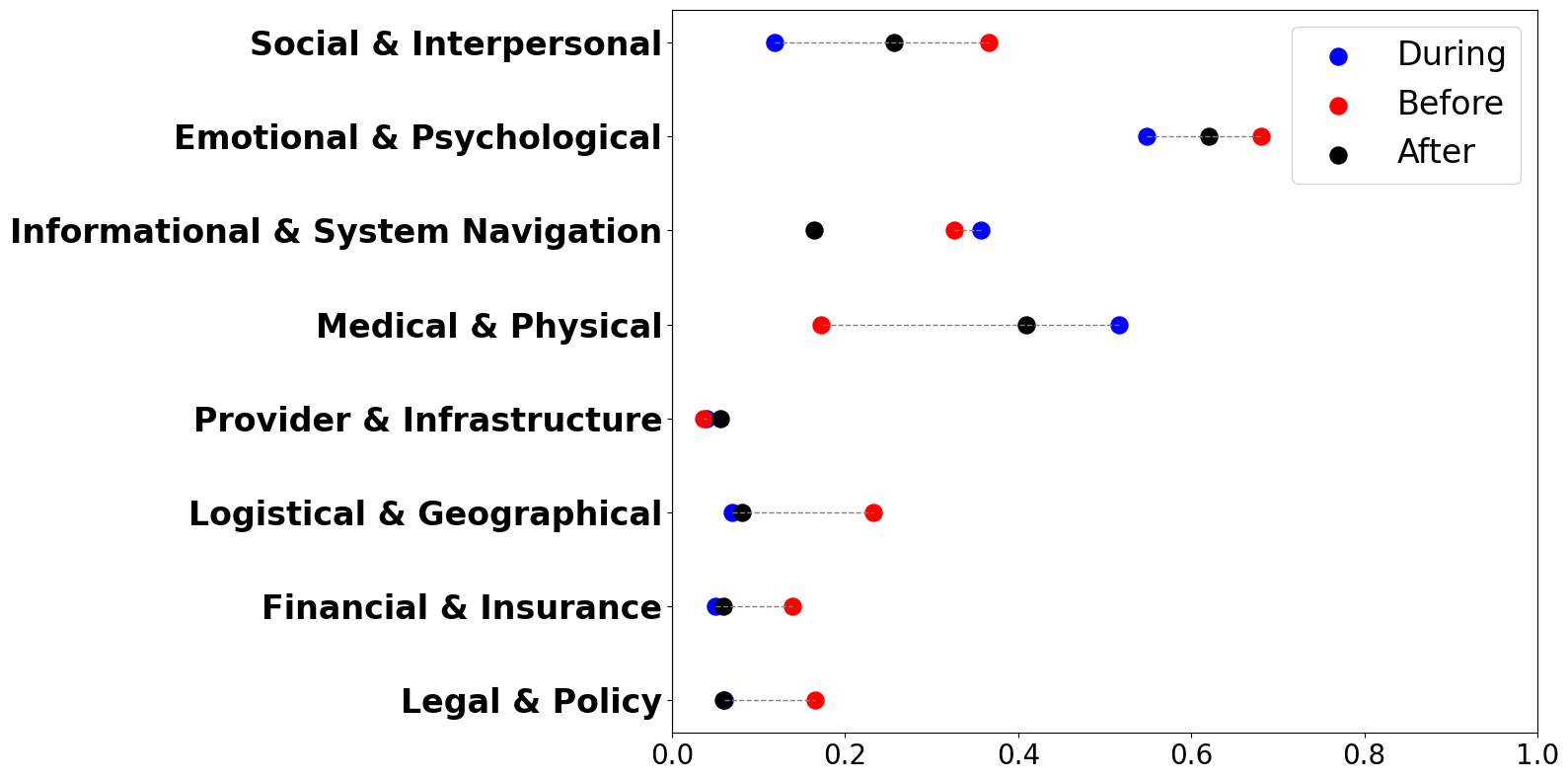}
    \caption{Barriers in different abortion stages}
    \label{fig:Barriers_Stages}
\end{subfigure}

\vspace{0.5em}

\begin{subfigure}[t]{\columnwidth}
    \centering
    \includegraphics[width=0.8\columnwidth]{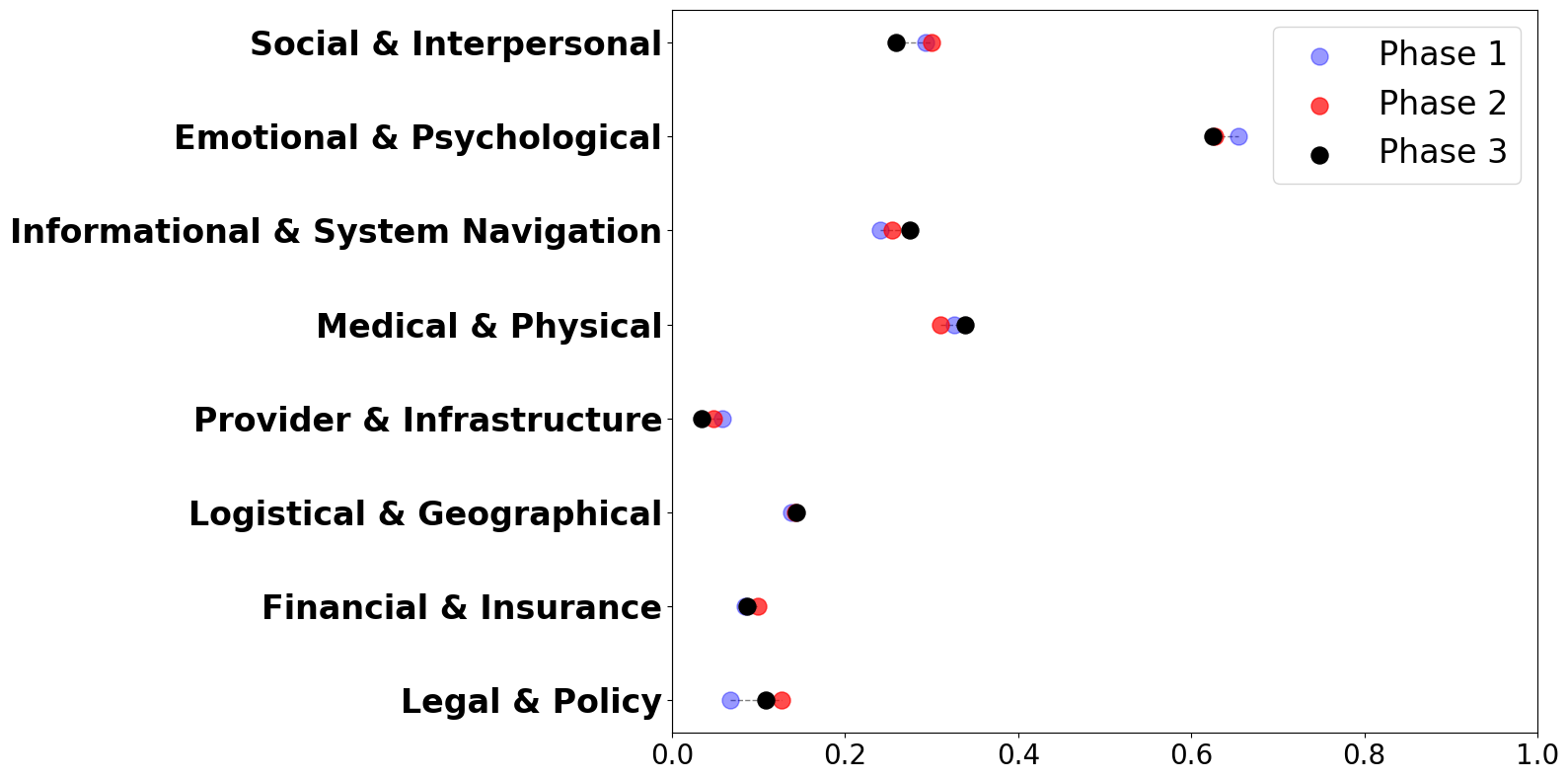}
    \caption{Barriers in different phases of Dobbs}
    \label{fig:Barriers_Phases}
\end{subfigure}

\caption{Barriers across (a) information types, (b) stages of abortion, and (c) phases of Dobbs}
\label{fig:graph-subplots}
\vspace{-0.3cm}
\end{figure}

Figure \ref{fig:graph-subplots} presents the normalized distribution of posts across the eight barrier types within (a) information types, (b) stages of abortion, and (c) phases of the Dobbs decision. As shown in Figure \ref{fig:Barriers_Info}, posts tagged as sharing information discussed experiences related to \textit{Emotional \& Psychological} and \textit{Social \& Interpersonal} barriers, whereas information-seeking posts more frequently raised \textit{Informational} and \textit{Medical} barriers. 
Other barrier types were distributed relatively evenly across the two information behaviors.
Figure \ref{fig:Barriers_Stages} shows that most barrier types were mentioned more frequently \textit{Before} abortion. 
By contrast, \textit{Medical} and \textit{Informational} barriers were most prominent \textit{During} abortion, while \textit{Emotional \& Psychological} barriers remain most prevalent across all three stages.
Figure \ref{fig:Barriers_Phases} indicates no substantial differences in barrier prevalence across the three phases of the Dobbs timeline. 

\begin{table*}[t]
\centering
\resizebox{0.7\textwidth}{!}{
\begin{tabular}{l c c c c c c c c c }
\toprule
Model & \begin{tabular}[c]{@{}l@{}}LP\end{tabular} & \begin{tabular}[c]{@{}l@{}}FI\end{tabular} & \begin{tabular}[c]{@{}l@{}}LG\end{tabular} & \begin{tabular}[c]{@{}l@{}}PI\end{tabular} & \begin{tabular}[c]{@{}l@{}}MP\end{tabular} & \begin{tabular}[c]{@{}l@{}}IS\end{tabular} & \begin{tabular}[c]{@{}l@{}}EP\end{tabular} & \begin{tabular}[c]{@{}l@{}} SI \end{tabular} & Average F1 \\
\hline
GPT 4 & 0.88 & 0.61 & 0.81 & 0.67 & 0.56 & 0.62 & 0.85 & 0.84 & 0.73 \\
GPT 4.1 & 0.79 & 0.72 & 0.79 & 0.38 & 0.58 & 0.57 & 0.75 & 0.73 & 0.66 \\
\textbf{GPT 4.1-mini} & 0.81 & 0.69 & 0.81 & 0.55 & 0.57 & 0.64 & 0.8 & 0.86 & \textbf{0.72} \\
GPT 4.1-nano & 0.76 & 0.76 & 0.77 & 0.36 & 0.57 & 0.43 & 0.85 & 0.84 & 0.67 \\
\textbf{GPT 4o} & 0.92 & 0.8 & 0.82 & 0.67 & 0.58 & 0.7 & 0.8 & 0.72 & \textbf{0.75} \\
GPT 4o-mini & 0.79 & 0.59 & 0.77 & 0.44 & 0.55 & 0.72 & 0.8 & 0.82 & 0.68 \\
\textbf{GPT 5.1} & 0.81 & 0.78 & 0.83 & 0.6 & 0.65 & 0.7 & 0.78 & 0.77 & \textbf{0.74} \\
GPT 5.2 & 0.8 & 0.6 & 0.68 & 0.29 & 0.56 & 0.56 & 0.74 & 0.74 & 0.62 \\
Llama-4-Maverick & 0.85 & 0.54 & 0.7 & 0.5 & 0.59 & 0.59 & 0.76 & 0.76 & 0.66 \\
Mixtral-8x7B & 0.6 & 0.12 & 0.47 & 0.57 & 0.19 & 0.43 & 0.54 & 0.72 & 0.45  \\
\hline
Ensemble (Top 3 Models\\ excluding GPT 4) & 0.85 & 0.75 & 0.84 & 0.67 & 0.59 & 0.74 & 0.8 & 0.8 & \textbf{0.75} \\ \hline
\end{tabular}}
\caption{Performance of different LLMs on 8 Barrier Types. LP: Legal \& Policy, FI: Financial \& Insurance, LG: Logistical \& Geographical, PI: Provider \& Infrastructure, MP: Medical \& Physical, IS: Informational \& System Navigation, EP: Emotional \& Psychological, SI: Social \& Interpersonal.}
\label{tab:Barriers Classification Models}
\end{table*}

\subsection{RQ3: Emotion Analysis}
Our analysis showed \textit{nervousness}, \textit{confusion}, \textit{fear}, \textit{sadness}, and \textit{curiosity} as the most salient emotions in our dataset (Figure \ref{fig:Most frequent emotions} in the Appendix), all closely tied to the uncertainty and sensitivity surrounding abortion discourse. Less frequent but still prominent emotions such as \textit{relief}, \textit{grief}, and \textit{caring} complemented these dominant categories, reflecting the complex emotional landscape of abortion-related narratives. Additionally, we found emotions such as \textit{nervousness} and \textit{confusion}, \textit{grief} and \textit{sadness}, and \textit{curiosity} and \textit{confusion} tend to co-occur within the same posts, highlighting the layered and multifaceted affective expressions in this discourse (Figure \ref{fig:Most frequent pairs of emotions that co-occur} in the Appendix).
Analyzing emotions across information types showed that posts classified as information-seeking were more likely to convey \textit{nervousness}, \textit{confusion}, \textit{curiosity}, and \textit{fear}, reflecting users' uncertainty and need for guidance (Figure \ref{fig:Emotions: DIfferential of Info Types} in the Appendix). By contrast, information-sharing posts more frequently expressed \textit{sadness}, \textit{relief}, and \textit{grief}, highlighting the role of sharing as a means of processing experiences and articulating personal outcomes. 

\begin{figure}[t]
    \centering
    
    \begin{subfigure}[t]{0.5\textwidth}
        \centering
        \includegraphics[width=0.8\columnwidth]{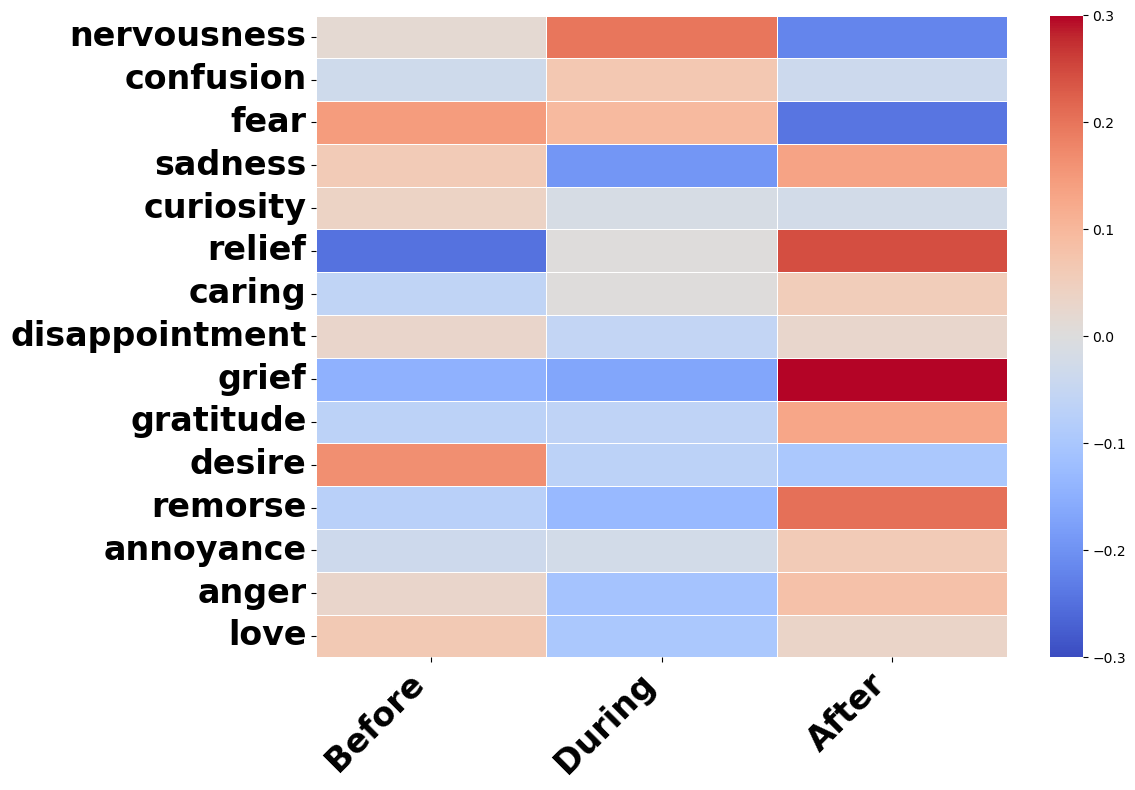}
        \caption{Emotions in Different Abortion Stages}
        \label{fig:Differential emotions in stages}
    \end{subfigure}
    \hfill

    \begin{subfigure}[t]{0.5\textwidth}
        \centering
        \includegraphics[width=0.8\columnwidth]{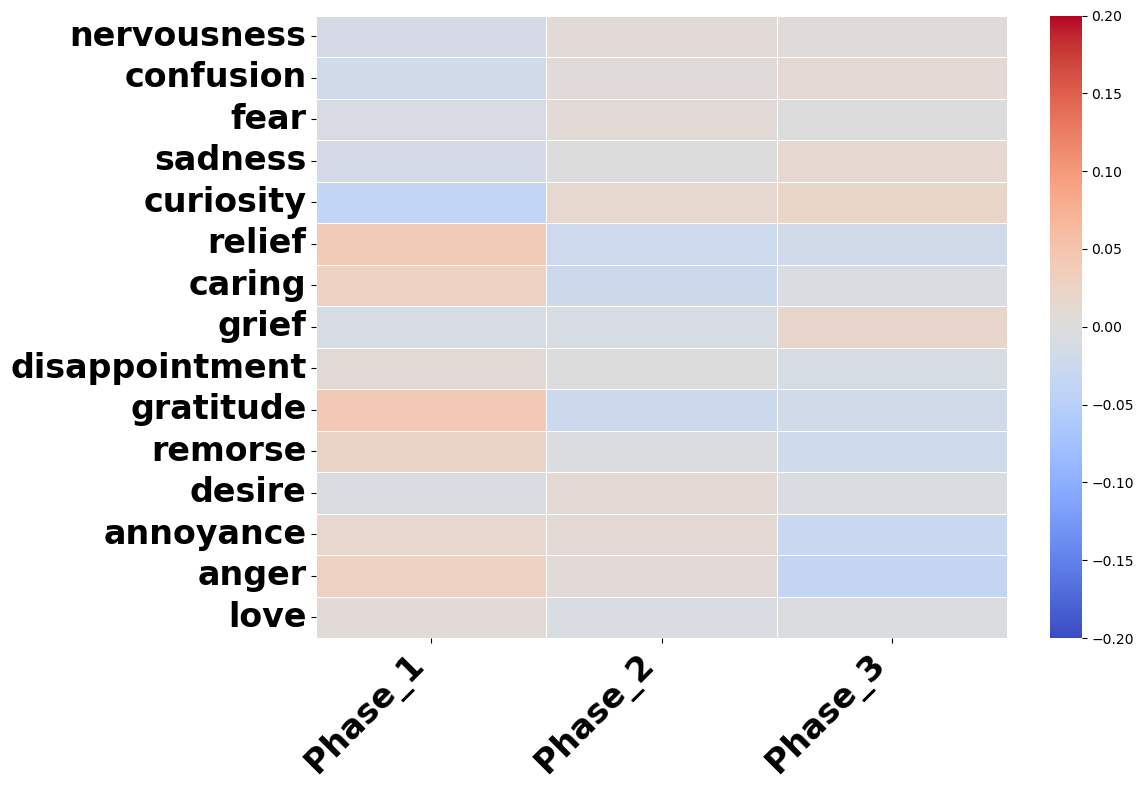}
        \caption{Emotions in Different Dobbs Phases}
        \label{fig: Differential emotions in phases}
    \end{subfigure}
    \hfill

    \caption{Emotions statistically stronger in: (a) Stages of abortion, (b) Phases of Dobbs}
    \label{fig:Differential emotions}
    \vspace{-0.8cm}
\end{figure}

\begin{figure}[t]
    \centering
        \includegraphics[width=0.8\columnwidth]{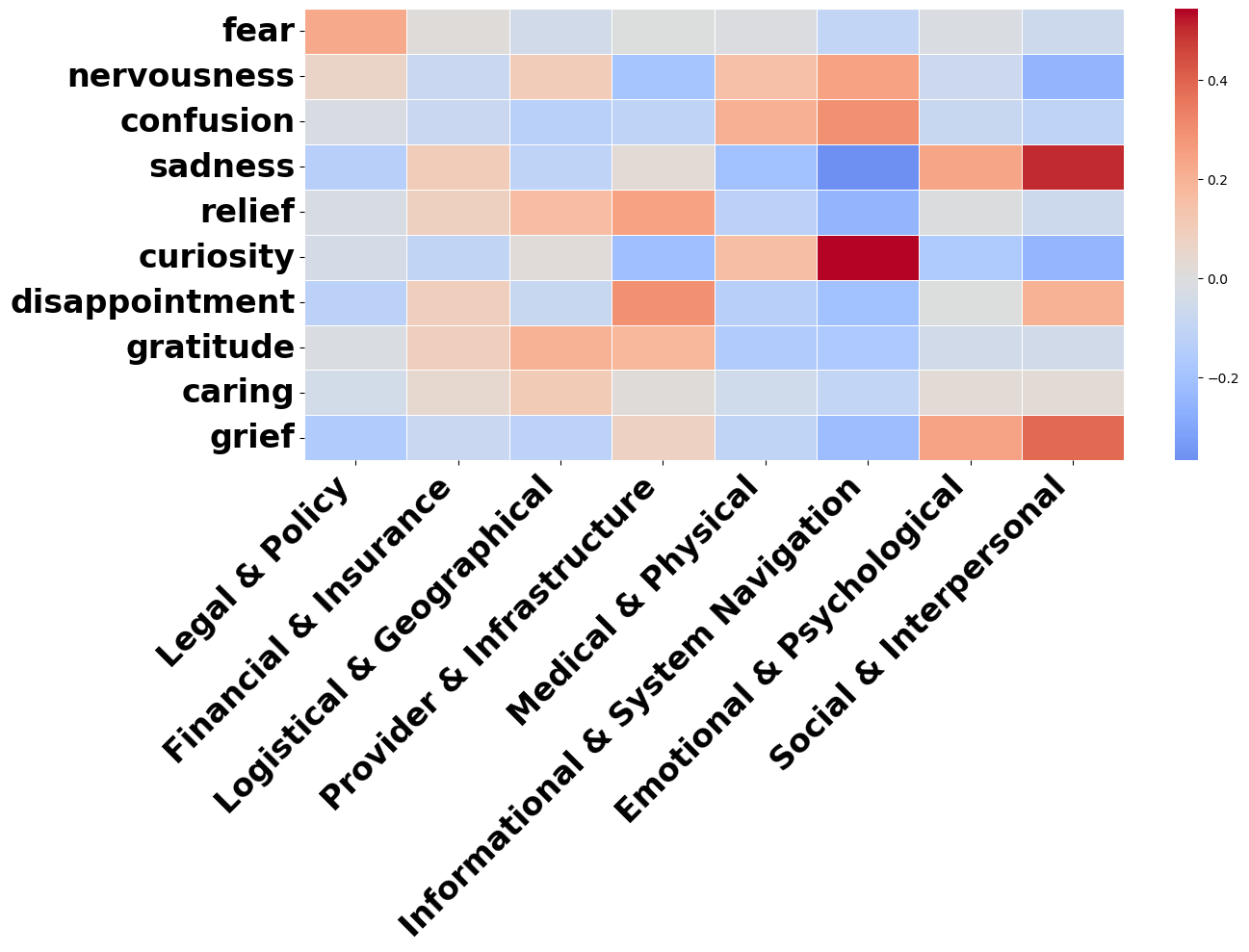}
    \caption{Emotions expressed with different barrier types}
    \label{fig:Emotions expressed related to different barriers of abortion}
\end{figure}

Clear distinctions emerged for expressed emotions across stages of abortion (Figure \ref{fig:Differential emotions in stages}): \textit{desire} and \textit{fear} appeared more frequently \textit{Before} abortion, \textit{nervousness} peaked \textit{During} abortion, while \textit{grief}, \textit{remorse}, and \textit{relief} were most prevalent \textit{After} abortion. 
Across three phases of the Dobbs decision (Figure \ref{fig: Differential emotions in phases}), emotions did not shift very strongly. 
Across barrier types, \textit{sadness} and \textit{grief} most frequently co-occurred with \textit{Social \& Interpersonal} barriers, \textit{curiosity} was most commonly associated with \textit{Informational \& System Navigation} barriers, and \textit{disappointment} was most prevalent in relation to \textit{Provider \& Infrastructure} barriers (Figure \ref{fig:Emotions expressed related to different barriers of abortion}). 

\subsection{Statistical Analysis Across Groups}
To assess whether the prevalence of barrier types and emotional expressions differed across three stages of abortion, three phases of Dobbs, and two information behaviors, we conducted a series of chi-square ($\chi^2$) tests. For each barrier category, we compared its distribution across the stages, Dobbs phases, and information behaviors to determine whether observed differences exceeded what would be expected by chance. We applied the same procedure when comparing emotion prevalence across these dimensions. Significance thresholds were set at $\alpha = 0.05$, and p-values are reported where relevant.

\noindent \textbf{Barriers Significance Tests Across Three Dimensions.}
Our results indicate that barriers vary more strongly with information behavior and abortion stage than with the phase of the Dobbs decision. By information type, the strongest differences emerged for \textit{Emotional \& Psychological} ($\chi^2$=635.6, $p=3e^{-140}$), \textit{Informational \& System Navigation} ($\chi^2$=525.1, $p=3e^{-116}$), and \textit{Social \& Interpersonal} ($\chi^2$=442.5, $p=3e^{-98}$). 
For abortion stages, \textit{Medical \& Physical} barriers ($\chi^2$=694.8, $p=e^{-151}$) and \textit{Logistical \& Geographical} barriers ($\chi^2$=390.5, $p=e^{-85}$) were found more significant. 
By contrast, phase effects were marginal; only \textit{Legal \& Policy} ($p=e^{-11}$), \textit{Provider \& Infrastructure} ($p=e^{-4}$), and \textit{Social \& Interpersonal} ($p=9e^{-4}$) vary across phases; implying Dobbs may have slightly changed people's worries (especially legal uncertainty), but the dominant drivers of barriers were independent of the political time window. Table \ref{tab:chi2_barriers} in the Appendix shows the complete results of the tests.

\noindent \textbf{Emotions Significance Tests Across Three Dimensions.}
Results showed that emotional expression was strongly shaped by information behavior, moderately shaped by abortion stage, and largely stable across Dobbs phases. Across information type, all prevalent emotions exhibited large and highly significant differences, with particularly strong effects for \textit{nervousness} ($\chi^2$= 1136.68, $p=3.5e^{-249}$), relief ($\chi^2$= 1075.6, $p=6.7e^{-236}$), \textit{confusion} ($\chi^2$= 958, $p=2.4e^{-210}$), and \textit{curiosity} ($\chi^2$= 873.89, $p=4.7e^{-192}$), suggesting that seeking or sharing information corresponded to a major shift in emotional tone. Across abortion stages, emotions also differed significantly, particularly for \textit{fear} ($\chi^2$= 540.8, $p=3.6e^{-118}$), \textit{relief} ($\chi^2$= 502.2, $p=8.9e^{-110}$), \textit{nervousness} ($\chi^2$= 452.4, $p=5.7e^{-99}$), and \textit{grief} ($\chi^2$= 413.5, $p=1.6e^{-90}$), reinforcing that emotional expression tracked experiential progression through the abortion journey. By contrast, emotions showed minimal phase sensitivity, with no meaningful Dobbs phase effects for most emotions, and even the few statistically significant patterns, e.g., \textit{curiosity} ($\chi^2$= 7.3, $p=2.6e^{-2}$) and \textit{relief} ($\chi^2$= 6.2, $p=4.6e^{-2}$) remained extremely small in magnitude. Overall, these results support a consistent interpretation that Dobbs may have shifted the intensity and urgency of discussion, but the emotional landscape was driven primarily by informational behavior and abortion stage, not by broad temporal phases around the Court decision. Table \ref{tab:chi2_emotions} in the Appendix shows the full results.

\subsection{Topic Modeling for Barriers}

\subsubsection{Barriers Across Phases. } Table \ref{tab:Topic Barrier Phase} in Appendix shows that across the three phases, the abortion-related barriers discussed on Reddit remained broadly consistent, but shifted in emphasis and specificity. In Phase 1, topics in posts tagged as \textit{Legal \& Policy} barriers centered on early pregnancy uncertainty, informal or underground pill access, and timing/appointment constraints. Financial barriers focused on cost anxiety, insurance/payment concerns, and legitimacy issues around ordering pills online. \textit{Logistical} barriers highlighted clinic scheduling and coordinating abortion timing around work and daily life. Posts in \textit{Medical} barriers emphasized Medical Abortion (MA) symptom management, testing confusion after abortion, and comparisons between Surgical Abortion (SA) and MA experiences. \textit{Informational \& System Navigation} barriers reflected difficulties navigating services and understanding medication instructions. \textit{Emotional \& Psychological} barriers foregrounded ambivalence, fear, guilt, and difficult decision-making, while social barriers emphasized disclosure dilemmas, relationship pressure, isolation, and support needs.
In Phase 2, approaching the Dobbs ruling, the discourse became more explicitly resource- and access-oriented. \textit{Legal \& Policy} barriers shifted toward pregnancy discovery, state restrictions, and navigating aid resources. \textit{Financial} discussions increasingly reflected urgent financial help-seeking, online ordering/payment issues, and hardship shaping abortion decisions. \textit{Logistical} barriers became more operational, focusing on travel and restrictive-state constraints as well as decision uncertainty under time pressure. \textit{Medical} discourse centered on bleeding duration and physical distress, while informational challenges emphasized accessing medical services, misoprostol usage confusion, and customs-related delays. \textit{Emotional \& Psychological} narratives largely condensed into emotional overload during uncertainty and decision fatigue, and \textit{Social \& Interpersonal} barriers emphasized partner dynamics and privacy/secrecy concerns.
By Phase 3, the post-Dobbs environment appears to further intensify execution and monitoring concerns tied to self-managed access pathways. Discourse related to \textit{Legal} barriers emphasized legal risk anxiety, cross-region pill access, and legality/week limits. \textit{Financial} barriers shifted toward general financial stress, access inequality, and support needs. \textit{Logistical} discourse was dominated by shipping/tracking issues and resource seeking in restrictive settings. \textit{Medical} discourse became more detailed, highlighting bleeding/clot duration, and side effects or infection concerns. Informational confusion similarly became more granular, focusing on detailed timing questions, symptom monitoring, and verification or follow-up care. Emotionally, decision conflict and distress remained prominent, while social themes emphasized support, disclosure strategies, and managing conversations under work and privacy constraints.

\subsubsection{Barriers Across Info Types. }
Across barrier types, distinct differences emerged between information types as shown in Table \ref{tab:Topic Barrier Info} in the Appendix. \textit{Legal} barriers illustrate this contrast, where people more distinctly seek guidance on legality, restrictions, timing, and access to abortion pills, highlighting confusion about laws and procedures. Similarly, \textit{Financial} barriers showed that information-seeking was more specifically tailored, while sharing themes were more generic. The divergence was most pronounced in \textit{Medical \& Physical} barriers. Seeking conversations were expansive, covering a wide range of anxieties from side effects, and cramping to infection concerns and prolonged bleeding—indicating a demand for reassurance and medical clarity. Shared discussions, on the other hand, contained broader narratives of emotional experiences, effectiveness concerns, and physical symptoms, often framed in personal reflection rather than detailed inquiries. \textit{Informational} barrier narratives followed this pattern as well, with seekers focusing on uncertainty around instructions, services, and medication use, while sharers focused on generic medication experiences.
\textit{Emotional \& Psychological} barriers also highlighted that seekers emphasized isolation, regret, fear, and uncertainty, whereas sharers more often disclosed lingering emotional impacts such as coping with depression, shame, or guilt. Finally, in social domains, seekers voiced concerns about a lack of support and partner influence, while sharing reflected lived struggles with stigma and inadequate partner support.

\section{Discussion}
\label{sec:discussion}

Our study demonstrates how Reddit functions as a reproductive health infrastructure in the wake of the Dobbs decision, where individuals not only debate abortion as a public issue, but also navigate barriers of abortion as a lived and time-sensitive experience through information exchange and emotional expression. By operationalizing abortion discourse through three intersecting dimensions: \textit{information behaviors}, \textit{stages of abortion}, and \textit{phases of Dobbs decision}, we provide a multi-dimensional account of how barriers to abortion care and its underlying emotions shape the narratives of abortion at scale. Our framework extends prior work showing that abortion-related online spaces combine informational needs and socio-emotional support \cite{john2024abortion, pleasants2024abortion, valdez2024analyzing}, by showing that these needs are not mentioned in isolation, but rather are interconnected and multi-faceted. Our work contributes (1) a large-scale computational analysis of abortion-related barriers and emotions following a major policy shock, (2) a methodological framework that jointly models information behavior, experiential stage, and political context, and (3) a conceptual insight that information behavior and abortion stage structure discourse more strongly than policy timing alone.

\noindent\textbf{Reddit as a Dual Infrastructure. }
A key implication of our framework is that information seeking and information sharing function as distinct social infrastructures rather than mere stylistic variation. Rooted in information behavior scholarship \cite{poltrock2003information, savolainen2005everyday} and online health community research \cite{de2014mental}, this distinction clarifies what the platform does for abortion-related needs. Seeking posts play a significant role in actionable decision-making, especially through system navigation and \textit{Medical \& Physical} concerns, suggesting Reddit's functionality as an informal coordination layer when formal health systems are difficult to access. This mirrors findings from harm-reduction and care-seeking research showing how people construct alternative pathways of support under conditions of institutional constraint \cite{gillespie2018custodians}. Sharing posts, by contrast, primarily reflect \textit{Emotional} and \textit{Social} constraints, aligning with the platform's role as a space for disclosure, coping, and peer validation, consistent with prior evidence that abortion forums provide both logistical advice and socio-emotional support \cite{wilson2024seeking, pleasants2024normal}.
Our statistical tests indicate that these behavioral differences are substantive: barrier prevalence and emotional expression vary most strongly across information behaviors. This contributes a computational complement to qualitative accounts of abortion support networks by demonstrating systematic differences between navigating structural uncertainty and processing emotional challenges. The prominence of information seeking \textit{Before} abortion highlights the need for clearer, accessible, and safety-oriented guidance in online spaces, particularly around system navigation and medical uncertainty \cite{pleasants2024normal, john2024abortion}. At the same time, the prevalence of information sharing \textit{After} abortion underscores the importance of preserving supportive environments for disclosure, emotional processing, and peer validation \cite{kimport2011social}. Platforms hosting abortion-related discussions may benefit from design interventions such as resource pinning, clearer pathways to verified medical information, and moderation strategies that balance harm reduction with privacy and anonymity in legally sensitive cases \cite{gillespie2020content}.

\noindent\textbf{Barriers and Emotions as Compounded Constraints. }
Across categories, the prominence and co-occurrence of \textit{Emotional \& Psychological} and \textit{Social \& Interpersonal} barriers underscore that abortion access is widely experienced and narrated as a compounded burden rather than a single challenge. While structural factors such as cost, travel, provider availability, and legal restrictions are well-established barriers to abortion access \cite{culwell2013addressing, doran2015barriers, higgins2021real}, our results indicate that Reddit narratives are dominated by affective and relational constraints, showing that abortion is often impacted by social circumstances and psychological distress \cite{norris2011abortion, hanschmidt2016abortion}. 
The co-occurrence patterns suggest that emotional burdens frequently appear alongside other barrier types, especially social barriers. 
This aligns with reproductive health research emphasizing that abortion-related distress and hardship are shaped by interacting constraints rather than isolated factors \cite{upadhyay2012contraceptive, finer2016declines}.

Our analysis of emotions expressed in the abortion discourse highlights the prevalence of emotions that correspond to uncertainty (\textit{confusion}, \textit{nervousness}, \textit{curiosity}) and vulnerability (\textit{fear}, \textit{sadness}), mirroring prior evidence that people seek abortion information to manage procedural and legal ambiguity \cite{john2024abortion}. The stage-based differences further support abortion as a temporal experience with shifting psychological demands \cite{jerman2017barriers, kimport2011social} where emotions track progression through the abortion journey, suggesting that Reddit serves as a continued coping space. Critically, the tight links between informational barriers and \textit{curiosity}, and between social barriers and \textit{sadness} and \textit{grief}, illustrate how institutional complexity and relational challenges intersect with emotionally charged experiences.

\noindent\textbf{Dobbs as an Amplifier of Discourse. }
The discourse volume increased around the Dobbs ruling and remains elevated afterward, consistent with prior work showing that major political events trigger spikes in online engagement and reframing \cite{chang2023roeoverturned, venkata2024post}. However, when we examine the discourse using normalized distributions and chi-square tests, most barrier and emotion categories show marginal differences across phases, suggesting that the relative structure of reported barriers is broadly stable over time. Where phase-level differences do appear, they are concentrated in a small subset of barrier types, e.g., \textit{Legal \& Policy}, \textit{Provider \& Infrastructure}, and \textit{Social \& Interpersonal}), and these effects are modest in magnitude compared to the much stronger differences observed across information behaviors and abortion stages. These findings suggest that the Dobbs decision coincided with increased Reddit engagement but only limited shifts in the structure of abortion narratives, with dominant patterns in barriers and emotions shaped more strongly by information behavior and abortion stage than by policy timing alone. At the same time, the limited phase-level differences suggest that while discussion volume increased following the policy change, the underlying types of barriers people face remained largely unchanged. This could indicate that core challenges persist regardless of shifts in the legal landscape. Further analysis is needed to interpret phase effects as well as the long-term impacts of these policy changes on personal barriers faced by abortion seekers. Our results reinforce the finding that barriers and emotions are shaped more by the experiential stage and information behavior than by policy timing alone.

\section{Limitations}
\label{sec:limitations}
Our work has several limitations: First, our classification of Reddit posts into abortion stages, barriers, and emotions relies in part on large language models. Although we validated model outputs against human-annotated data and conducted reliability checks, misclassification and model bias remain possible and may affect the precision of some results.
Second, Reddit is not representative of all abortion discourse. The data analyzed here reflects English-speaking, publicly articulated user-selected experiences shaped by the norms and affordances of the platform. As a result, our findings characterize patterns in online abortion discourse rather than the prevalence or distribution of experiences in the broader population, and should be understood as complementary to clinical, survey-based, and qualitative research.
Third, our dataset covers a one-year window surrounding the Dobbs decision. While this period captures immediate and short-term responses, the longer-term social, legal, and emotional consequences of such policy changes may continue to evolve beyond the time frame studied here. Finally, statistical analysis relied primarily on chi-square tests, which capture pairwise associations but do not account for higher-order interactions across variables; modeling such interactions was beyond the scope of this study. Additionally, the interpretation of results emphasizes statistical significance, and the absence of effect size measures limits the ability to assess the magnitude of observed relationships.

\section{Conclusion}
\label{sec:conclusion}
Our analysis of more than 17K abortion-related Reddit posts shows that while information seeking often reflects on informational barriers and associated feelings like curiosity, confusion, and fear, information sharing highly emphasizes emotional and social barriers, and feelings of sadness, relief, and disappointment. 
Our barrier framework demonstrates how legal, financial, logistical, and medical obstacles are deeply entangled with emotional and social burdens such as stigma, secrecy, and lack of support. While \textit{nervousness} is the most prevalent emotion overall, \textit{Social} conflicts are highly linked to \textit{sadness}, and \textit{Informational} barriers are tied to \textit{curiosity}. The findings of topic modeling indicate the multi-layered, complex nature of abortion across two information types and three phases of Dobbs. However, emotions and barriers are evenly distributed between these phases.

\section{Acknowledgement}
We are grateful to Kelsey Chong and Kristine Yoo for their assistance with data preparation. We also thank OpenAI for providing the research credits used to conduct the experiments. Finally, we would like to acknowledge all the Reddit users whose discussions made this work possible.
\label{sec:ack}

\bibliography{aaai26}

\newpage

\subsection{Paper Checklist:}

\begin{enumerate}

\item For most authors...
\begin{enumerate}
    \item  Would answering this research question advance science without violating social contracts, such as violating privacy norms, perpetuating unfair profiling, exacerbating the socio-economic divide, or implying disrespect to societies or cultures?
    \answerTODO{Yes, our contributions are laid out in Introduction section}
  \item Do your main claims in the abstract and introduction accurately reflect the paper's contributions and scope?
    \answerTODO{Yes- Added in the Discussion section}
   \item Do you clarify how the proposed methodological approach is appropriate for the claims made? 
    \answerTODO{Yes- Added in the Discussion section}
   \item Do you clarify what are possible artifacts in the data used, given population-specific distributions?
    \answerTODO{NA}
  \item Did you describe the limitations of your work?
    \answerTODO{Yes- in the Limitation section}
  \item Did you discuss any potential negative societal impacts of your work?
    \answerTODO{NA- no potential negative impact}
      \item Did you discuss any potential misuse of your work?
    \answerTODO{NA- no potential misuse}
    \item Did you describe steps taken to prevent or mitigate potential negative outcomes of the research, such as data and model documentation, data anonymization, responsible release, access control, and the reproducibility of findings?
    \answerTODO{Yes- data is anonymized, and will be shared later}
  \item Have you read the ethics review guidelines and ensured that your paper conforms to them?
    \answerTODO{Yes}
\end{enumerate}

\item Additionally, if your study involves hypotheses testing...
\begin{enumerate}
  \item Did you clearly state the assumptions underlying all theoretical results?
    \answerTODO{NA}
  \item Have you provided justifications for all theoretical results?
    \answerTODO{NA}
  \item Did you discuss competing hypotheses or theories that might challenge or complement your theoretical results?
    \answerTODO{NA}
  \item Have you considered alternative mechanisms or explanations that might account for the same outcomes observed in your study?
    \answerTODO{NA}
  \item Did you address potential biases or limitations in your theoretical framework?
    \answerTODO{NA}
  \item Have you related your theoretical results to the existing literature in social science?
    \answerTODO{NA}
  \item Did you discuss the implications of your theoretical results for policy, practice, or further research in the social science domain?
    \answerTODO{NA}
\end{enumerate}

\item Additionally, if you are including theoretical proofs...
\begin{enumerate}
  \item Did you state the full set of assumptions of all theoretical results?
    \answerTODO{NA}
	\item Did you include complete proofs of all theoretical results?
    \answerTODO{NA}
\end{enumerate}

\item Additionally, if you ran machine learning experiments...
\begin{enumerate}
  \item Did you include the code, data, and instructions needed to reproduce the main experimental results (either in the supplemental material or as a URL)?
    \answerTODO{No, they will be added to the final version}
  \item Did you specify all the training details (e.g., data splits, hyperparameters, how they were chosen)?
    \answerTODO{Yes- in the Method section}
     \item Did you report error bars (e.g., with respect to the random seed after running experiments multiple times)?
    \answerTODO{NA}
	\item Did you include the total amount of compute and the type of resources used (e.g., type of GPUs, internal cluster, or cloud provider)?
    \answerTODO{NA- we did not use GPUs or cloud}
     \item Do you justify how the proposed evaluation is sufficient and appropriate to the claims made? 
    \answerTODO{Yes- in the Discussion section}
     \item Do you discuss what is ``the cost`` of misclassification and fault (in)tolerance?
    \answerTODO{Yes- we mentioned misclassificatio impacts our results in the Limitation section}
  
\end{enumerate}

\item Additionally, if you are using existing assets (e.g., code, data, models) or curating/releasing new assets, \textbf{without compromising anonymity}...
\begin{enumerate}
  \item If your work uses existing assets, did you cite the creators?
    \answerTODO{Yes- in the Method Section}
  \item Did you mention the license of the assets?
    \answerTODO{Yes- we cited the tools}
  \item Did you include any new assets in the supplemental material or as a URL?
    \answerTODO{No}
  \item Did you discuss whether and how consent was obtained from people whose data you're using/curating?
    \answerTODO{NA- our data is anonymized}
  \item Did you discuss whether the data you are using/curating contains personally identifiable information or offensive content?
    \answerTODO{NA- our data is anonymized}
\item If you are curating or releasing new datasets, did you discuss how you intend to make your datasets FAIR (see \citet{wilkinson2016fair})?
\answerTODO{No- the data will be shared later with proper documentation}
\item If you are curating or releasing new datasets, did you create a Datasheet for the Dataset (see \citet{gebru2021datasheets})? 
\answerTODO{No-we have reported some high-level characteristics of the dataset}
\end{enumerate}

\item Additionally, if you used crowdsourcing or conducted research with human subjects, \textbf{without compromising anonymity}...
\begin{enumerate}
  \item Did you include the full text of instructions given to participants and screenshots?
    \answerTODO{NA}
  \item Did you describe any potential participant risks, with mentions of Institutional Review Board (IRB) approvals?
    \answerTODO{NA}
  \item Did you include the estimated hourly wage paid to participants and the total amount spent on participant compensation?
    \answerTODO{NA}
   \item Did you discuss how data is stored, shared, and deidentified?
   \answerTODO{NA}
\end{enumerate}

\end{enumerate}

\bigskip

\appendix

\section{Appendix}
\label{sec:app}

\subsection{Co-occurring Barriers}

Table \ref{tab:Barriers that co-occur} shows the most frequent pairs of barrier types that co-occur in the same posts. As we see, emotional \& Psychological frequently comes alongside other barrier types, with its co-occurrence with Social \& Interpersonal barrier in 25 percent of all posts. This confirms that barriers to reproductive care not only has structural, societal aspects, but it also affectively impacts individuals at the same time. We also notice that medical and informational barriers, and social and financial challenges come together in a number of posts, suggesting the multi-faceted nature of experiencing barriers to abortion. 

\begin{table}[h]
\centering
\begin{tabular}{lc}
\toprule
Pair     & Normalized No. \\
\hline
EP \& SI & 0.25           \\
EP \& MP & 0.11           \\
EP \& IS & 0.07           \\
EP \& LG & 0.07           \\
EP \& FI & 0.06           \\
MP \& IS & 0.05           \\
EP \& LP & 0.04           \\
SI \& FI & 0.04           \\
SI \& LG & 0.03           \\
LP \& LG & 0.03          \\\hline
\end{tabular}
\caption{Normalized frequency of barrier types that co-occur in posts.  LP: Legal \& Policy, FI: Financial \& Insurance, LG: Logistical \&
Geographical, PI: Provider \& Infrastructure, MP: Medical \& Physical, IS: Informational \& System Navigation, EP: Emotional
\& Psychological, SI: Social \& Interpersonal}
\label{tab:Barriers that co-occur}
\end{table}

\subsection{Validation of Emotion Analysis}
To validate the credibility of Go Emotions in our dataset domain, we applied our Emotion Classification Prompt on a random sample of 500 texts from the training dataset in Go Emotions paper \cite{demszky2020goemotions} and compared LLM prediction using GPT-4o (same model applied to our abortion dataset) with ground-truth labels in this data that included 28 emotions. Our result showed that the LLM zero-shot classification achieved $0.62$ accuracy, suggesting the overall adaptability and applicability of this framework for our study.

\subsection{Most Frequent Emotions}
\begin{figure}[t]
    \centering
        \includegraphics[width=\columnwidth]{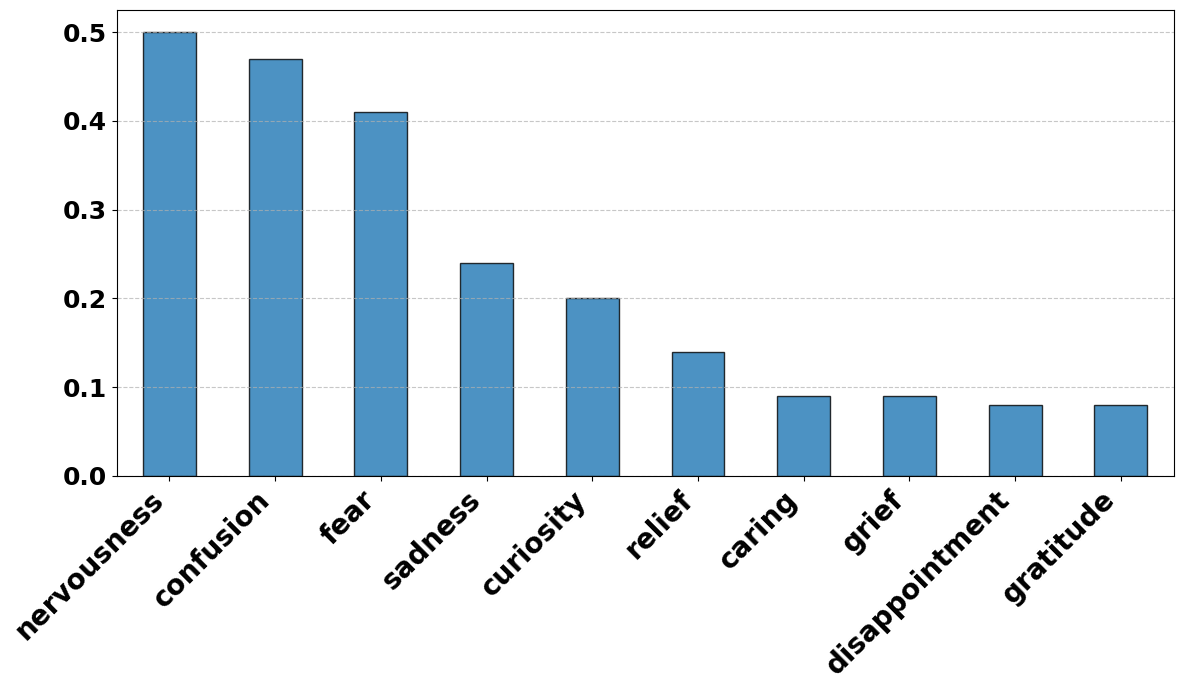}
    \caption{Most frequent emotions}
    \label{fig:Most frequent emotions}
\end{figure}

\begin{figure}[t]

        \centering
        \includegraphics[width=\columnwidth]{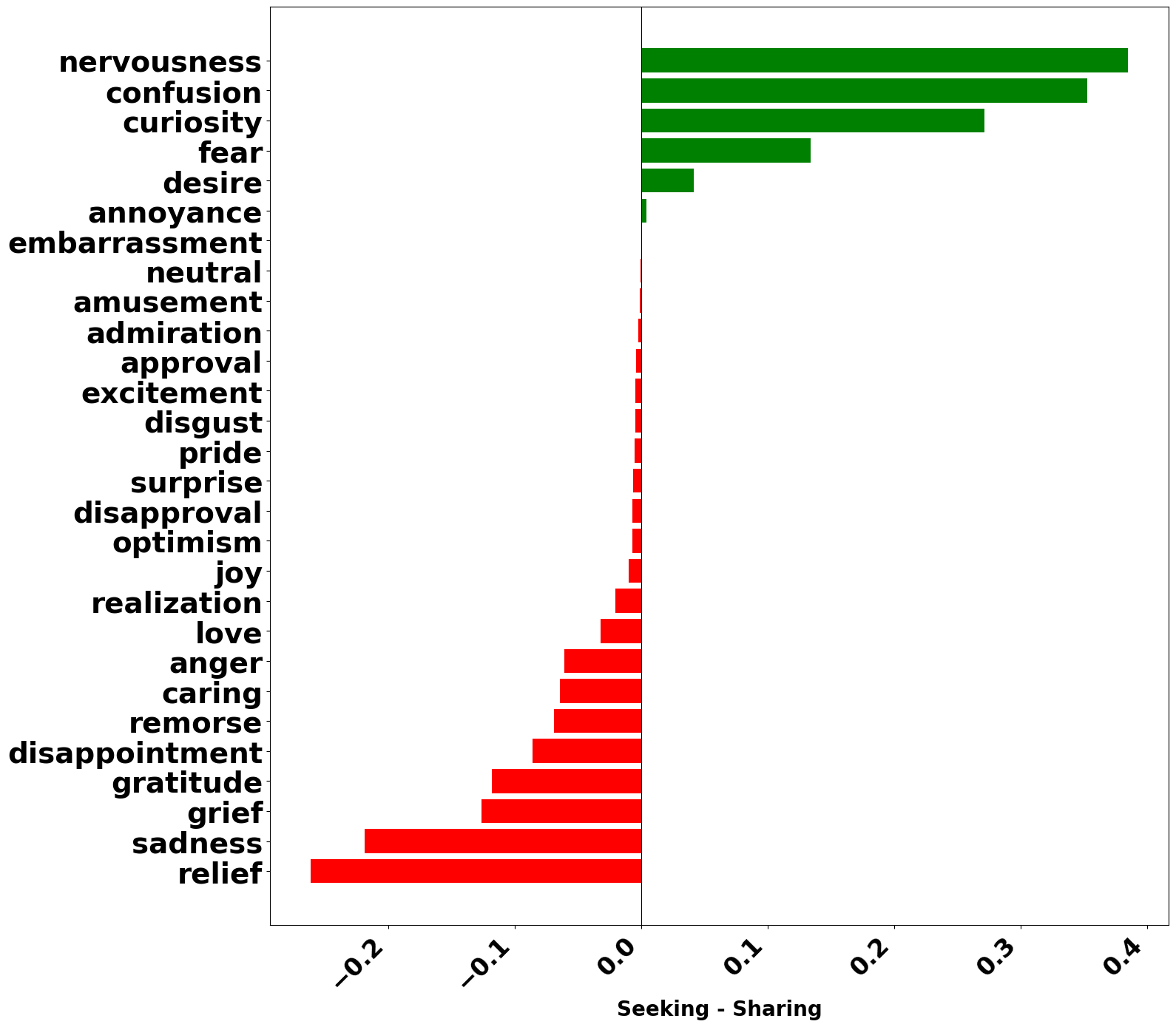}
        \caption{Emotion in Seeking vs. Sharing. Positive values indicate seeking is more frequent, negative values reflect more frequent sharing.}
        \label{fig:Emotions: DIfferential of Info Types}
\end{figure}

Figure \ref{fig:Most frequent emotions} shows the most frequent emotions expressed in the data predicted by GPT-4.1-mini based on Go Emotions framework. \textit{Fear}, \textit{Sadness}, \textit{Confusion} and \textit{Nervousness} are most prominent expressed emotions in our abortion dataset.
Figure \ref{fig:Emotions: DIfferential of Info Types} shows emotions in information sharing vs. sharing posts.

\subsection{Co-occurring Emotions}
\begin{figure}[t]
        \includegraphics[width=0.5\textwidth]{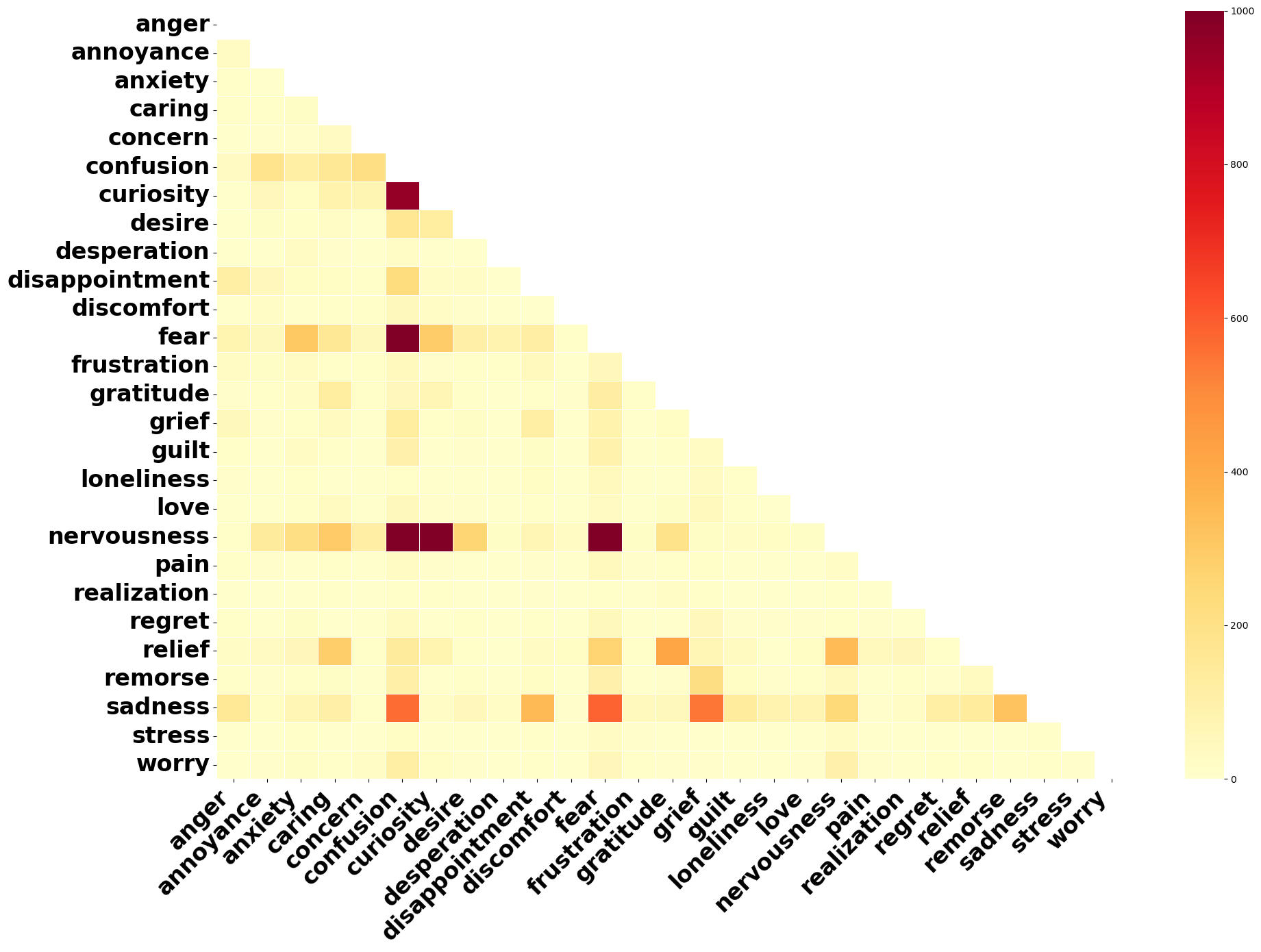}

    \caption{Most frequent pairs of emotions that co-occur}
    \label{fig:Most frequent pairs of emotions that co-occur}
\end{figure}

Figure \ref{fig:Most frequent pairs of emotions that co-occur} shows the heatmap of how frequently different pairs of emotions co-occur in a same posts. As we can see, `Fear' appears frequently with `Nervousness', `Confusion' and `Sadness', showing the strong association of various negative feelings in the context of abortion. Similarly, `Nervousness' appears a lot with `Curiosity' and `Confusion', further confirming that restrictions and uncertainty in reproductive care affect vulnerable individuals seeking credible sources of information. It is insightful to see that`Relief', an emotion mostly expressed after the abortion is co-occurring occasionally with other positive emotions such as `caring' and `Gratitude', hinting at a positive sentiment after the care is received.

\subsection{LLM Prompts}
We designed multiple zero-shot prompts to use LLMs for different classification tasks. For detecting the stage of abortion, we used (Prompt: Abortion Stage Classification). For identifying multiple emotions in text, we designed (Prompt: Abortion Stage Classification). Finally, in order to task the model with detecting appropriate barrier types in text based on our framework, we used (Prompt: Abortion Barrier Classification) below.

\begin{table}[t]
\centering
\small
\setlength{\tabcolsep}{8pt}
\renewcommand{\arraystretch}{1.15}

\begin{tabular}{lr}
\toprule
\textbf{Model} & \textbf{F1 Score} \\
\midrule
GPT-4.1      & 0.91 \\
GPT-4.1-mini & 0.87 \\
GPT-5-mini   & 0.87 \\
GPT-5-nano   & 0.81 \\
\bottomrule
\end{tabular}
\caption{F1 scores for abortion stage classification across four LLMs evaluated on the same held-out labeled test set. Higher values indicate better agreement with human-annotated stage labels.}
\label{tab:stage_f1_models}
\end{table}

\begin{table*}[t]
\centering
\small
\setlength{\tabcolsep}{6pt}
\renewcommand{\arraystretch}{1.15}

\begin{tabular}{llrrrrl}
\toprule
\textbf{Group} & \textbf{Barrier} & \textbf{Statistic} & \textbf{p-value} & \textbf{Cram\'er's V} & \textbf{Significant} \\
\midrule
Information Type & Legal \& Policy & 0.82 & $3.66e^{-1}$ & 0.010 & No \\
Information Type & Financial \& Insurance & 14.33 & $1.53e^{-4}$ & 0.041 & Yes \\
Information Type & Logistical \& Geographical & 10.62 & $1.12e^{-3}$ & 0.036 & Yes \\
Information Type & Provider \& Infrastructure & 64.65 & $8.95e^{-16}$ & 0.088 & Yes \\
Information Type & Medical \& Physical & 92.81 & $5.75e^{-22}$ & 0.105 & Yes \\
Information Type & Informational \& System Navigation & 525.06 & $3.35e^{-116}$ & 0.251 & Yes \\
Information Type & Emotional \& Psychological & 635.58 & $3.06e^{-140}$ & 0.276 & Yes \\
Information Type & Social \& Interpersonal & 442.49 & $3.11e^{-98}$ & 0.230 & Yes \\
\midrule
Abortion Stage & Legal \& Policy & 244.80 & $6.94e^{-54}$ & 0.171 & Yes \\
Abortion Stage & Financial \& Insurance & 168.83 & $2.19e^{-37}$ & 0.142 & Yes \\
Abortion Stage & Logistical \& Geographical & 390.52 & $1.58e^{-85}$ & 0.216 & Yes \\
Abortion Stage & Provider \& Infrastructure & 16.48 & $2.63e^{-4}$ & 0.044 & Yes \\
Abortion Stage & Medical \& Physical & 694.77 & $1.36e^{-151}$ & 0.288 & Yes \\
Abortion Stage & Informational \& System Navigation & 300.13 & $6.71e^{-66}$ & 0.189 & Yes \\
Abortion Stage & Emotional \& Psychological & 62.90 & $2.20e^{-14}$ & 0.087 & Yes \\
Abortion Stage & Social \& Interpersonal & 282.92 & $3.68e^{-62}$ & 0.184 & Yes \\
\midrule
Phase & Legal \& Policy & 49.75 & $1.57e^{-11}$ & 0.077 & Yes \\
Phase & Financial \& Insurance & 4.05 & $1.32e^{-1}$ & 0.022 & No \\
Phase & Logistical \& Geographical & 0.40 & $8.17e^{-1}$ & 0.007 & No \\
Phase & Provider \& Infrastructure & 18.35 & $1.04e^{-4}$ & 0.047 & Yes \\
Phase & Medical \& Physical & 5.40 & $6.70e^{-2}$ & 0.025 & No \\
Phase & Informational \& System Navigation & 7.74 & $2.08e^{-2}$ & 0.030 & No \\
Phase & Emotional \& Psychological & 5.69 & $5.80e^{-2}$ & 0.026 & No \\
Phase & Social \& Interpersonal & 14.01 & $9.05e^{-4}$ & 0.041 & Yes \\
\bottomrule
\end{tabular}
\caption{Chi-square test results for barrier prevalence across information type, abortion stage, and Dobbs phase.}
\label{tab:chi2_barriers}
\end{table*}

\begin{table*}[t]
\centering
\small
\setlength{\tabcolsep}{6pt}
\renewcommand{\arraystretch}{1.15}

\begin{tabular}{llrrrrl}
\toprule
\textbf{Group} & \textbf{Emotion} & \textbf{Statistic} & \textbf{p-value} & \textbf{Cram\'er's V} & \textbf{Significant} \\
\midrule
Information Type & nervousness & 1136.68 & $3.53e^{-249}$ & 0.368648 & Yes \\
Information Type & confusion & 958.01 & $2.41e^{-210}$ & 0.338437 & Yes \\
Information Type & fear & 141.04 & $1.58e^{-32}$ & 0.129854 & Yes \\
Information Type & sadness & 509.40 & $8.57e^{-113}$ & 0.246787 & Yes \\
Information Type & curiosity & 873.89 & $4.66e^{-192}$ & 0.323237 & Yes \\
Information Type & relief & 1075.57 & $6.73e^{-236}$ & 0.358602 & Yes \\
Information Type & caring & 92.51 & $6.71e^{-22}$ & 0.105167 & Yes \\
Information Type & grief & 360.19 & $2.56e^{-80}$ & 0.207520 & Yes \\
Information Type & disappointment & 182.57 & $1.33e^{-41}$ & 0.147743 & Yes \\
Information Type & gratitude & 370.93 & $1.18e^{-82}$ & 0.210590 & Yes \\
\midrule
Abortion Stage & nervousness & 452.43 & $5.71e^{-99}$ & 0.232578 & Yes \\
Abortion Stage & confusion & 28.88 & $5.34e^{-7}$ & 0.058765 & Yes \\
Abortion Stage & fear & 540.85 & $3.60e^{-118}$ & 0.254291 & Yes \\
Abortion Stage & sadness & 101.72 & $8.17e^{-23}$ & 0.110279 & Yes \\
Abortion Stage & curiosity & 8.62 & $1.35e^{-2}$ & 0.032095 & No \\
Abortion Stage & relief & 502.19 & $8.93e^{-110}$ & 0.245034 & Yes \\
Abortion Stage & caring & 25.91 & $2.36e^{-6}$ & 0.055658 & Yes \\
Abortion Stage & grief & 413.49 & $1.63e^{-90}$ & 0.222345 & Yes \\
Abortion Stage & disappointment & 5.68 & $5.85e^{-2}$ & 0.026051 & No \\
Abortion Stage & gratitude & 76.61 & $2.31e^{-17}$ & 0.095707 & Yes \\
\midrule
Phase & nervousness & 2.58 & $2.75e^{-1}$ & 0.017562 & No \\
Phase & confusion & 3.92 & $1.41e^{-1}$ & 0.021645 & No \\
Phase & fear & 1.00 & $6.07e^{-1}$ & 0.010921 & No \\
Phase & sadness & 2.16 & $3.40e^{-1}$ & 0.016055 & No \\
Phase & curiosity & 7.27 & $2.64e^{-2}$ & 0.029487 & No \\
Phase & relief & 6.17 & $4.57e^{-2}$ & 0.027159 & No \\
Phase & caring & 3.57 & $1.68e^{-1}$ & 0.020666 & No \\
Phase & grief & 2.02 & $3.64e^{-1}$ & 0.015544 & No \\
Phase & disappointment & 0.39 & $8.22e^{-1}$ & 0.006842 & No \\
Phase & gratitude & 6.74 & $3.44e^{-2}$ & 0.028390 & No \\
\bottomrule
\end{tabular}
\caption{Chi-square test results for the top 10 emotions across information type, abortion stage, and Dobbs phase.}
\label{tab:chi2_emotions}
\end{table*}

\begin{table*}
 \centering
 \small
\begin{tabular}{cll}

\hline
                                                                                                &                         &                                                                                                                                                                                                                                                                                                         \\
\multirow{-2}{*}{Barrier Type}                                                                  & \multirow{-2}{*}{Phase} & \multirow{-2}{*}{Topics}                                                                                                                                                                                                                                                                                \\ \hline
                                                                                                & Phase 1                 & \cellcolor[HTML]{FFFFFF}{\color[HTML]{1F1F1F} Early Pregnancy uncertainty- Informal/Underground Pill access- MA Timing and Appointment Constraints }                                                                                                                                                                               \\ \cline{2-3} 
                                                                                                & Phase 2                 & \cellcolor[HTML]{FFFFFF}{\color[HTML]{1F1F1F} Pregnancy Discovery- State Restrictions- Navigating Aid resources}                                                                                                                                                                                     \\ \cline{2-3} 
\multirow{-3}{*}{Legal \& Policy}                                                               & Phase 3                 & \cellcolor[HTML]{FFFFFF}{\color[HTML]{1F1F1F} Legal Risk Anxiety- Cross-Region Pill Access- Legality and Weeks Limits}                                                                                                                                                                       \\ \hline
                                                                                                & Phase 1                 & \cellcolor[HTML]{FFFFFF}{\color[HTML]{1F1F1F} Cost Anxiety- Ordering Pills Online: Payment + Legitimacy- Financial barriers}                                                                                                                                                                                              \\ \cline{2-3} 
                                                                                                & Phase 2                 & \cellcolor[HTML]{FFFFFF}{\color[HTML]{1F1F1F} Financial Help-Online Ordering- Financial hardship decisions}                                                                                                                                                                                              \\ \cline{2-3} 
\multirow{-3}{*}{\begin{tabular}[c]{@{}c@{}}Financial \& \\ Insurance\end{tabular}}             & Phase 3                 & \cellcolor[HTML]{FFFFFF}{\color[HTML]{1F1F1F} \begin{tabular}[c]{@{}l@{}}General Financial Stress- Access Inequality- Financial support \end{tabular}}                                                                                 \\ \hline
                                                                                                & Phase 1                 & \cellcolor[HTML]{FFFFFF}{\color[HTML]{1F1F1F} Clinic Scheduling- Abortion Timing + Work/Life Constraints- Shipping/Arrival Delays}                                                                                                                                                                                                                               \\ \cline{2-3} 
                                                                                                & Phase 2                 & \cellcolor[HTML]{FFFFFF}{\color[HTML]{1F1F1F} Travel, Resources, and State Constraints- Decision Uncertainty Under Time Pressur}                                                                                                                                                                         \\ \cline{2-3} 
\multirow{-3}{*}{\begin{tabular}[c]{@{}c@{}}Logistical \& \\ Geographical\end{tabular}}         & Phase 3                 & \cellcolor[HTML]{FFFFFF}{\color[HTML]{1F1F1F} Shipping/Tracking- Resource Seeking for Restrictive States}                                                                                                                                                                             \\ \hline
                                                                                                & Phase 1                 & \cellcolor[HTML]{FFFFFF}{\color[HTML]{1F1F1F} Planned Parenthood / Clinic Scheduling Bottlenecks-Confusion and dissatisfaction tied to provider interactions}                                                                                                                                                                         \\ \cline{2-3} 
                                                                                                & Phase 2                 & \cellcolor[HTML]{FFFFFF}{\color[HTML]{1F1F1F} Provider Experience Narratives- provider access gaps}                                                                                                                                                                                                                                 \\ \cline{2-3} 
\multirow{-3}{*}{\begin{tabular}[c]{@{}c@{}}Provider \& \\ Infrastructure\end{tabular}}         & Phase 3                 & \cellcolor[HTML]{FFFFFF}{\color[HTML]{1F1F1F} Provider Communication as a Source of Confusion- Difficulty getting through the pipeline}                                                                                                                                                                                                       \\ \hline
                                                                                                & Phase 1                 & \cellcolor[HTML]{FFFFFF}{\color[HTML]{1F1F1F} \begin{tabular}[c]{@{}l@{}}MA Symptom Management- Testing Confusion After Abortion- SA vs. MA Experience\end{tabular}}                                     \\ \cline{2-3} 
                                                                                                & Phase 2                 & \cellcolor[HTML]{FFFFFF}{\color[HTML]{1F1F1F} Bleeding Duration- Physical Distress}                                                                                                                                                                             \\ \cline{2-3} 
\multirow{-3}{*}{\begin{tabular}[c]{@{}c@{}}Medical \& \\ Physical\end{tabular}}                & Phase 3                 & \cellcolor[HTML]{FFFFFF}{\color[HTML]{1F1F1F} \begin{tabular}[c]{@{}l@{}}Bleeding + Clots + Duration- Pain, Cramps, and Symptoms- Side Effects and Infection Concerns \end{tabular}}                                                                                        \\ \hline
                                                                                                & Phase 1                 & \cellcolor[HTML]{FFFFFF}{\color[HTML]{1F1F1F} \begin{tabular}[c]{@{}l@{}}Navigating services- Medication effectiveness concerns- Post-abortion symptoms- \\ Service access confusion- Abortion medication instructions- Medical decision anxiety\end{tabular}}                                      \\ \cline{2-3} 
                                                                                                & Phase 2                 & \cellcolor[HTML]{FFFFFF}{\color[HTML]{1F1F1F} \begin{tabular}[c]{@{}l@{}}Accessing medical services- Uncertainty and confusion- Misoprostol usage confusion- \\ Service access confusion- Customs clearance delays\end{tabular}}                                                                        \\ \cline{2-3} 
\multirow{-3}{*}{\begin{tabular}[c]{@{}c@{}}Informational \& \\ System Navigation\end{tabular}} & Phase 3                 & \cellcolor[HTML]{FFFFFF}{\color[HTML]{1F1F1F} \begin{tabular}[c]{@{}l@{}}Detailed timing question- Symptom Monitoring- Verification and Follow Up Care\end{tabular}} \\ \hline
                                                                                                & Phase 1                 & \cellcolor[HTML]{FFFFFF}{\color[HTML]{1F1F1F} \begin{tabular}[c]{@{}l@{}}Ambivalence and Moral Conflict About the Decision- Fear and Nervousness About Procedures- \\ Anxiety and panic- Coping with guilt- Difficult personal decisions- Emotional isolation stigma\end{tabular}}                      \\ \cline{2-3} 
                                                                                                & Phase 2                 & \cellcolor[HTML]{FFFFFF}{\color[HTML]{1F1F1F} \begin{tabular}[c]{@{}l@{}}Emotional Overload During Uncertainty- Decision Fatigue Under Time Pressure\end{tabular}}                                                                                           \\ \cline{2-3} 
\multirow{-3}{*}{\begin{tabular}[c]{@{}c@{}}Emotional \& \\ Psychological\end{tabular}}         & Phase 3                 & \cellcolor[HTML]{FFFFFF}{\color[HTML]{1F1F1F} \begin{tabular}[c]{@{}l@{}}Decision Conflict- Emotional distress- Fear of pain\end{tabular}}                                                   \\ \hline
                                                                                                & Phase 1                 & \cellcolor[HTML]{FFFFFF}{\color[HTML]{1F1F1F} \begin{tabular}[c]{@{}l@{}}Disclosure Dilemmas: Parents, Partners, Friends- Relational Conflict- \\ Isolation in Young/Dependent Contexts- Support Needs\end{tabular}}            \\ \cline{2-3} 
                                                                                                & Phase 2                 & \cellcolor[HTML]{FFFFFF}{\color[HTML]{1F1F1F} Partner Dynamics and Emotional Negotiation-Privacy and secrecy- validation and support}                                                                                                                                                                  \\ \cline{2-3} 
\multirow{-3}{*}{\begin{tabular}[c]{@{}c@{}}Social \& \\ Interpersonal\end{tabular}}            & Phase 3                 & \cellcolor[HTML]{FFFFFF}{\color[HTML]{1F1F1F} Telling family/partners, managing conversations- work/privacy constraints}                                                                                                                                                                                                        \\ \hline
\end{tabular}
\caption{Topics extracted for each Barrier in 3 Phases}
\label{tab:Topic Barrier Phase}
\end{table*}

\clearpage{}

\begin{table*}
\centering
\small
\begin{tabular}{cll}
\hline
                                                                                                &                             &                                                                                                                                                                                                                                                                                                                                                                                                                                                                                                                              \\
\multirow{-2}{*}{Barrier Type}                                                                  & \multirow{-2}{*}{Info Type} & \multirow{-2}{*}{Topics}                                                                                                                                                                                                                                                                                                                                                                                                                                                                                                     \\ \hline
                                                                                                & Seeking                     & \cellcolor[HTML]{FFFFFF}{\color[HTML]{1F1F1F} \begin{tabular}[c]{@{}l@{}}Abortion legality concerns- Abortion legal restrictions- Abortion access restrictions- \\ Accessing abortion pills- Parental consent abortion- Abortion timing laws- Customs and legal anxiety\end{tabular}}                                                                                                                                                                                                                                        \\ \cline{2-3} 
\multirow{-2}{*}{Legal \& Policy}                                                               & Sharing                     & \cellcolor[HTML]{FFFFFF}{\color[HTML]{1F1F1F} Abortion access barriers- Roe v. Wade}                                                                                                                                                                                                                                                                                                                                                                                                                                         \\ \hline
                                                                                                & Seeking                     & \cellcolor[HTML]{FFFFFF}{\color[HTML]{1F1F1F} Abortion access barriers- Financial hardship abortion- Abortion cost barriers}                                                                                                                                                                                                                                                                                                                                                                                                 \\ \cline{2-3} 
\multirow{-2}{*}{\begin{tabular}[c]{@{}c@{}}Financial \& \\ Insurance\end{tabular}}             & Sharing                     & \cellcolor[HTML]{FFFFFF}{\color[HTML]{1F1F1F} Abortion access barriers}                                                                                                                                                                                                                                                                                                                                                                                                                                                      \\ \hline
                                                                                                & Seeking                     & \cellcolor[HTML]{FFFFFF}{\color[HTML]{1F1F1F} \begin{tabular}[c]{@{}l@{}}Abortion access challenges- Delays accessing medication- Package customs delays- \\ Abortion pill access- Abortion appointment delays\end{tabular}}                                                                                                                                                                                                                                                                                                 \\ \cline{2-3} 
\multirow{-2}{*}{\begin{tabular}[c]{@{}c@{}}Logistical \& \\ Geographical\end{tabular}}         & Sharing                     & \cellcolor[HTML]{FFFFFF}{\color[HTML]{1F1F1F} Abortion access challenges- Shipping delays concerns- Abortion pill access}                                                                                                                                                                                                                                                                                                                                                                                                    \\ \hline
                                                                                                & Seeking                     & \cellcolor[HTML]{FFFFFF}{\color[HTML]{1F1F1F} Abortion care challenges- Abortion access trauma}                 \\ \cline{2-3} 
\multirow{-2}{*}{\begin{tabular}[c]{@{}c@{}}Provider \& \\ Infrastructure\end{tabular}}         & Sharing                     & \cellcolor[HTML]{FFFFFF}{\color[HTML]{1F1F1F} Abortion access barriers- Abortion care experiences}                                                                                                                                                                                                                                                                                                                                                                                                                           \\ \hline
                                                                                                & Seeking                     & \cellcolor[HTML]{FFFFFF}{\color[HTML]{1F1F1F} \begin{tabular}[c]{@{}l@{}}Medication abortion anxiety- Post-abortion bleeding- Nausea and vomiting-\\  Medical anxiety concerns- Heavy menstrual bleeding- Abortion side effects-\\ Severe post-abortion cramping- Nausea during procedures- Misoprostol effectiveness concerns- \\ Infection concerns- Abortion timing concerns- Prolonged bleeding post-abortion- \\ Health-related eligibility barriers- Misoprostol side effects- Persistent abdominal pain\end{tabular}} \\ \cline{2-3} 
\multirow{-2}{*}{\begin{tabular}[c]{@{}c@{}}Medical \& \\ Physical\end{tabular}}                & Sharing                     & \cellcolor[HTML]{FFFFFF}{\color[HTML]{1F1F1F} \begin{tabular}[c]{@{}l@{}}Abortion emotional experiences- Medication effectiveness anxiety- \\ Uterine health concerns- Abortion physical symptoms\end{tabular}}                                                                                                                                                                                                                                                                                                              \\ \hline
                                                                                                & Seeking                     & \cellcolor[HTML]{FFFFFF}{\color[HTML]{1F1F1F} \begin{tabular}[c]{@{}l@{}}Abortion information uncertainty- Healthcare access confusion-Accessing Services - \\ Medication instructions confusion- Pregnancy testing confusion-\\ Medication administration methods- Medication dosage confusion\end{tabular}} \\ 
\cline{2-3} 
\multirow{-2}{*}{\begin{tabular}[c]{@{}c@{}}Informational \& \\ System Navigation\end{tabular}} & Sharing                     & \cellcolor[HTML]{FFFFFF}{\color[HTML]{1F1F1F} Abortion medication experiences}                                                                                                                                                                                                                                                                                                                                                                                                                                               \\ \hline
                                                                                                & Seeking                     & \cellcolor[HTML]{FFFFFF}{\color[HTML]{1F1F1F} \begin{tabular}[c]{@{}l@{}}Emotional challenges abortion- Surgical procedure anxiety- Overwhelming negative emotions- \\ Isolation and Anxiety- Feeling Nervous Online- Regretful decision-making- Coping with Depression- \\ Abortion decision conflict- Early pregnancy anxiety- Fear of complications- Anxiety and fear-\end{tabular}}                                                                                                                                      \\ \cline{2-3} 
\multirow{-2}{*}{\begin{tabular}[c]{@{}c@{}}Emotional \& \\ Psychological\end{tabular}}         & Sharing                     & \cellcolor[HTML]{FFFFFF}{\color[HTML]{1F1F1F} \begin{tabular}[c]{@{}l@{}}Emotional impact abortion- Emotional support challenges- Fear and anxiety- Guilt and anxiety\\ Emotional aftermath of abortion-Coping with depression- Abortion anxiety experience- \\ Emotional coping struggles- Shame and guilt- Abortion decision regret- Pregnancy anxiety\end{tabular}}                                                                                                                                                       \\ \hline
                                                                                                & Seeking                     & \cellcolor[HTML]{FFFFFF}{\color[HTML]{1F1F1F} \begin{tabular}[c]{@{}l@{}}Lack of partner support- Partner-influenced abortion decisions- Relationship struggles post-abortion- \\ Abortion-related stigma- Lack of paternal support-\end{tabular}}                     \\ \cline{2-3} 
\multirow{-2}{*}{\begin{tabular}[c]{@{}c@{}}Social \& \\ Interpersonal\end{tabular}}            & Sharing                     & \cellcolor[HTML]{FFFFFF}{\color[HTML]{1F1F1F} Relationship support struggles- Abortion stigma experiences- Abortion stigma emotions-}                                                                                                                                                                                                                                                                                                                                                                                        \\ \hline
\end{tabular}
\caption{Topics extracted for each barrier in 2 Information Types}
\label{tab:Topic Barrier Info}
\end{table*}

\clearpage

\begin{tcolorbox}[colback=gray!5,colframe=black!80,title=Prompt: Abortion Stage Classification, label={box:stages of abortion}, width=0.5\textwidth]
You are an expert at identifying and detecting whether the text is related to a narrative, experience, or a situation before or after abortion.  

Read the following text and decide if it is related to a narrative, experience, or a situation before the abortion procedure is completed, or after it.  

Use one of the following labels:  

\begin{itemize}
  \item \textbf{After}: If the narrative, experience, situation discussed in the text happens after the abortion is completed.
  \item \textbf{Before}: If the narrative, experience, situation discussed in the text happens before the abortion is completed.
  \item \textbf{During}: If the narrative, experience, situation discussed in the text happens right in the middle of, or during the abortion (e.g., taking the first pill but waiting for the next).
  \item \textbf{Not Sure}: If the stage of abortion cannot be confidently identified.
  \item \textbf{Irrelevant}: If the text is about ideological debates (e.g., pro-choice vs pro-life), unrelated topics, news stories, or third-person accounts of another person’s abortion.
\end{itemize}

Use the following format for your response:  

\begin{verbatim}
[Label]
\end{verbatim}

Strictly follow this format. Do not include any additional text, commentary, or explanation.  

Now read the following text and provide the output:
\label{prompt1}
\end{tcolorbox}

\begin{tcolorbox}[colback=gray!5, colframe=black!80,
                  title=Prompt: Emotion Classification,
                  boxsep=1pt, width=0.5\textwidth]
\small

You are an expert emotion classification model.  
Your task is to read a given text and classify it into the most relevant emotions from the predefined list below.  

\textbf{You must follow these rules:}
\begin{itemize}
  \item Select up to 3 emotions (can be 1, 2, or 3) that best capture the emotional content of the text. 
  \item Always choose from the list provided. 
  \item Do not invent new categories. 
  \item If more than 3 emotions apply, select the ones most strongly implied. 
\end{itemize}

\textbf{Here is the list of available emotions with their brief definitions:}

\begin{enumerate}\setlength\itemsep{0.3em}
  \item[0.] admiration: Finding something impressive or worthy of respect  
  \item[1.] amusement: Finding something funny or being entertained  
  \item[2.] anger: A strong feeling of displeasure or antagonism  
  \item[3.] annoyance: Mild anger, irritation  
  \item[4.] approval: Having or expressing a favorable opinion  
  \item[5.] caring: Displaying kindness and concern for others  
  \item[6.] confusion: Lack of understanding, uncertainty  
  \item[7.] curiosity: A strong desire to know or learn something  
  \item[8.] desire: A strong feeling of wanting something or wishing for something to happen  
  \item[9.] disappointment: Sadness or displeasure caused by the nonfulfillment of one’s hopes or expectations  
  \item[10.] disapproval: Having or expressing an unfavorable opinion  
  \item[11.] disgust: Revulsion or strong disapproval aroused by something unpleasant or offensive  
  \item[12.] embarrassment: Self-consciousness, shame, or awkwardness  
  \item[13.] excitement: Feeling of great enthusiasm and eagerness  
  \item[14.] fear: Being afraid or worried  
  \item[15.] gratitude: A feeling of thankfulness and appreciation  
  \item[16.] grief: Intense sorrow, especially caused by someone’s death  
  \item[17.] joy: A feeling of pleasure and happiness  
  \item[18.] love: A strong positive emotion of regard and affection  
  \item[19.] nervousness: Apprehension, worry, anxiety  
  \item[20.] optimism: Hopefulness and confidence about the future or the success of something  
  \item[21.] pride: Pleasure or satisfaction due to one’s own achievements or those of close associates  
  \item[22.] realization: Becoming aware of something  
  \item[23.] relief: Reassurance and relaxation following release from anxiety or distress  
  \item[24.] remorse: Regret or guilty feeling  
  \item[25.] sadness: Emotional pain, sorrow  
  \item[26.] surprise: Feeling astonished, startled by something unexpected  
  \item[27.] neutral
\end{enumerate}

\textbf{Strictly follow this format:}  
\begin{verbatim}
[emotion1, emotion2, emotion3]
\end{verbatim}

Now, classify the following text:

\end{tcolorbox}


\begin{tcolorbox}[colback=gray!5,colframe=black!80,title=Prompt: Abortion Barrier Classification,label={box:barriers of abortion}, width=\textwidth]
\small
You are an advanced annotation assistant trained to identify barriers to abortion access.  
Your task is to read a given text and classify it into one or more of barrier categories.  
Additionally, you should explain the reasoning or mention the part of the text that made you tag that barrier type(s).

\textbf{Barrier Categories:}
\begin{enumerate}
  \item \textbf{Legal \& Policy Barriers} → Restrictions from laws, regulations, or policies that limit abortion access. 
   Includes: restrictive legislation, policies prohibiting abortions, prosecution or criminalization of abortion.\
   
  \item \textbf{Financial \& Insurance Barriers} → Obstacles tied to the cost of abortion care and lack of financial support.
   Includes: high out-of-pocket costs, related expenses (travel, childcare, lost wages), lack of insurance coverage, not being able to afford the abortion process.
  \item \textbf{Logistical \& Geographical Barriers} → Practical obstacles that make it difficult to obtain abortion care.
   Includes: long travel distances, lack of transportation, geographical limitations to access to abortion, appointment delays, time-sensitive access problems.
  \item \textbf{Provider \& Infrastructure Barriers} → Barriers rooted in the healthcare system or provider behavior.
   Includes: unsupportive provider attitudes, lack of professional medical support, shortage of providers, limited clinic infrastructure.
   
  \item \textbf{Medical \& Physical Barriers} → Health-related concerns or conditions that interfere with abortion access.
   Includes: existing health conditions, fear of medical risks, disability or chronic illness making travel difficult, emergency health needs left unmet.
  \item \textbf{Informational \& System Navigation Barriers} → Barriers that stem from lack of knowledge or helpful resources e.g., lack of reproductive-related knowledge, difficulty understanding healthcare systems, unhelpful or unclear guidance from healthcare professionals
  
  \item \textbf{Emotional \& Psychological Barriers} → Internal affective challenges that hinder access. 
   Includes: personal fear from abortion, anxiety, guilt, shame, distress about decision-making, mental health burdens like depression or trauma. 
  \item \textbf{Social \& Interpersonal Barriers} → Barriers related to the societal, familial or community-based pressures. 
   Includes: stigma, lack of partner/family support;  need for secrecy; secrecy due to fear of judgment; relationship conflict; social isolation; cultural or religious norms against abortion.  
\end{enumerate}

\textbf{Instructions:}
\begin{itemize}
  \item Identify barriers that are \textbf{explicitly described or strongly implied}.  
  \item Select up to 3 categories. If none apply, use \textbf{NA}.  
  \item Add \textbf{Other} if you detect a barrier outside the 8 types, with a brief explanation.  
  \item Focus only on barriers to abortion in \textbf{personal narratives}, not hypothetical political/ideological debates.
\end{itemize}

\textbf{Response Format:}
\begin{verbatim}
Barriers: [Label 1, Label 2, ...]
Cues: [Cue 1, Cue 2, ...]

If no Barriers:
Barriers: [NA]
Cues: [NA]
\end{verbatim}

Strictly follow this format. Do not include any additional text, commentary, or explanation.  

Now read the following text and provide the output:
\end{tcolorbox}

\end{document}